\documentclass[10pt,twocolumn,letterpaper]{article}

\usepackage[pagenumbers]{wacv} 

\usepackage{footmisc}
\usepackage{graphicx}
\usepackage{subcaption}
\usepackage{rotating}
\usepackage[export]{adjustbox}
\usepackage[table]{xcolor}

\usepackage{multirow}

\usepackage{algorithm}
\usepackage{algorithmic}
\usepackage{amsmath}
\usepackage{amssymb}
\usepackage{tikz}

\usepackage{booktabs, array, colortbl, amssymb}
\usepackage{rotating}
\usepackage{amsmath}
\usepackage{pifont}
\usepackage{dashrule}
\usepackage[table]{xcolor}

\definecolor{wacvblue}{rgb}{0.21,0.49,0.74}
\usepackage[pagebackref,breaklinks,colorlinks,allcolors=wacvblue]{hyperref}
\hypersetup{
    citecolor=wacvblue,       
    linkcolor=red,       
    urlcolor=blue          
}

\def\wacvPaperID{2003} 
\def\confName{WACV}
\def\confYear{2026}

\title{TIDE: Two-Stage Inverse Degradation Estimation with Guided Prior Disentanglement for Underwater Image Restoration}
\author{
    Shravan Venkatraman$^{1*}$ \quad
    Rakesh Raj Madavan$^{2*}$ \quad
    Pavan Kumar S$^{3*}$ \quad
    Muthu Subash Kavitha$^{4*}$ \\
    $^1$Mohamed bin Zayed University of AI \quad
    $^2$University of Amsterdam \\
    $^3$University of Massachusetts, Amherst \quad
    $^4$Nagasaki University
}

\begin{document}
\twocolumn[{%
    \renewcommand\twocolumn[1][]{#1}%
    \maketitle

    \begin{center}
        \captionsetup{type=figure}
        \includegraphics[width=\linewidth]{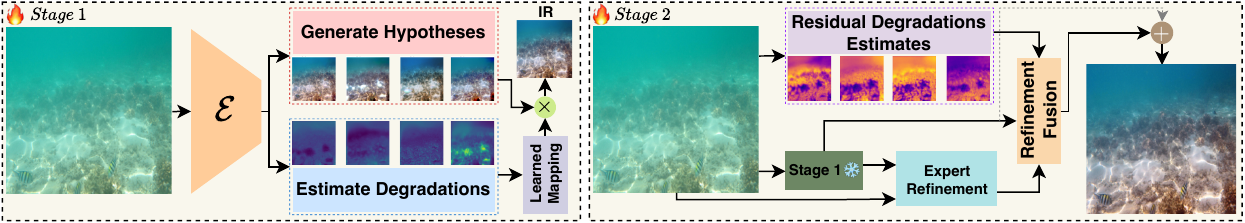}
        \caption{\textbf{Overview of the TIDE framework.} The pipeline consists of two stages. (1) Base restoration (left): The degraded image is processed by an encoder $\mathcal{E}$, producing features for degradation estimation and hypothesis generation. Multiple specialized restoration hypotheses are generated and fused based on estimated degradation maps via a learned mapping function. (2) Progressive refinement (right): The initial restoration and input image are analyzed to estimate residual degradations. Targeted corrections from refinement experts are selectively applied through a safety-gated fusion mechanism, yielding the final restored result.}
        \label{fig:teaser}
    \end{center}
}]

{\renewcommand{\thefootnote}{*}\footnotetext{Equal contribution.}}

\begin{abstract}
Underwater image restoration is essential for marine applications ranging from ecological monitoring to archaeological surveys, but effectively addressing the complex and spatially varying nature of underwater degradations remains a challenge. Existing methods typically apply uniform restoration strategies across the entire image, struggling to handle multiple co-occurring degradations that vary spatially and with water conditions. We introduce TIDE, a $\underline{t}$wo stage $\underline{i}$nverse $\underline{d}$egradation $\underline{e}$stimation framework that explicitly models degradation characteristics and applies targeted restoration through specialized prior decomposition. Our approach disentangles the restoration process into multiple specialized hypotheses that are adaptively fused based on local degradation patterns, followed by a progressive refinement stage that corrects residual artifacts. Specifically, TIDE decomposes underwater degradations into four key factors, namely color distortion, haze, detail loss, and noise, and designs restoration experts specialized for each. By generating specialized restoration hypotheses, TIDE balances competing degradation factors and produces natural results even in highly degraded regions. Extensive experiments across both standard benchmarks and challenging turbid water conditions show that TIDE achieves competitive performance on reference based fidelity metrics while outperforming state of the art methods on non reference perceptual quality metrics, with strong improvements in color correction and contrast enhancement. Our code is available at: \textcolor{cyan}{\url{https://rakesh-123-cryp.github.io/TIDE/}}.
\end{abstract}    
\section{Introduction}
\label{sec:intro}

Underwater image restoration (UIR) is a critical problem in computer vision with applications in marine biology, archaeology, robotics, and ocean engineering. Underwater images suffer from wavelength-dependent color distortion, low contrast, haze from scattering, blurriness, and noise, which degrade both visual quality and downstream tasks such as detection and scene understanding. Deep learning methods, including generative approaches~\cite{ugan,watergan}, zero-shot methods~\cite{zs1,zs2,zs3}, and medium-guided strategies~\cite{li2021underwater}, have advanced the field, yet they struggle with the complex, spatially varying nature of underwater degradations. A single image may simultaneously contain regions with different types and severities of distortions due to variations in depth, lighting, and water properties. However, existing methods implicitly assume a generic and homogeneous degradation distribution across images in a dataset, overlooking this inherent spatial and contextual variability. As a result, they often deliver suboptimal outcomes: when tuned for severely degraded regions, they over-process well-preserved areas and introduce artifacts; when calibrated conservatively, they leave challenging regions under-restored.

Recent approaches have attempted to address spatially varying degradations in underwater imagery. Multi-scale architectures~\cite{ms1,ms2_litnet,ms3,ms4} extract features at different resolutions to capture degradations of varying scales. Attention-based methods~\cite{fusion,ms2_litnet,att1,att2} dynamically weight features based on their importance for restoration. Some approaches~\cite{db1,db2,db3,db4} employ dual-branch networks that separately handle color correction and dehazing before fusion. Graph-based methods~\cite{graph1,graph2} model relationships between different image regions to adapt restoration parameters locally. While these approaches improve upon uniform processing, they typically generate a single restoration output directly and lack explicit mechanisms to handle the co-occurrence of multiple degradation types in different regions. Additionally, they often struggle with error accumulation, where inaccuracies in initial processing stages propagate to the final output without corrections.

To this end, we introduce TIDE, a framework for UIR through explicit degradation modeling and progressive refinement. TIDE implements two key innovations: (1) inverse degradation mapping with specialized prior decomposition, which disentangles different degradation types and applies targeted restoration strategies; and (2) two-stage degradation compensation, which identifies and corrects residual degradations after initial restoration. In the first stage, TIDE estimates pixel-wise degradation maps for color distortion, contrast reduction, detail loss, and noise. Then it generates specialized restoration hypotheses using decoders designed with inductive biases for each type of degradation. These hypotheses are dynamically fused based on the estimated degradation maps, producing a spatially varying restoration that addresses each region's specific degradation characteristics. Despite this targeted approach, severely degraded regions (e.g., turbid water or low-light areas) may retain residual artifacts after the first stage. Our second stage explicitly analyzes these residual degradations, generating targeted correction terms through focused refinement experts, and applying these corrections selectively to regions requiring further enhancement.

Figure~\ref{fig:teaser} illustrates the TIDE framework architecture. The base model first extracts multi-scale features from the degraded input image. These features serve two parallel paths: (1) degradation estimation, which produces pixel-wise maps indicating the spatial distribution and severity of different degradation types, and (2) hypothesis generation, where specialized decoders generate restoration candidates targeting specific degradation types. Each decoder embeds domain-specific inductive biases, and a learned mapping function adaptively fuses these hypotheses based on the degradation maps, producing an initial restoration. The second stage focuses on residual degradations by contrasting the input with the initial restoration. This differential signal steers refinement experts that predict targeted corrections, which are then fused through a gated mechanism to enhance only the degraded regions while preserving well-restored areas. Such progressive refinement lets TIDE handle complex, non-uniform degradations more effectively than single-stage methods, as seen in Figure~\ref{fig:teaser}.

To summarize, our main contributions are:

\begin{itemize}
\item We present TIDE, a framework that addresses complex, spatially varying underwater degradations through a structured, multi-stage restoration process.
\item We introduce a degradation-specific hypothesis generation and fusion strategy, which produces targeted corrections for distinct degradation types and combines them adaptively to handle heterogeneous distortions.
\item We develop a residual-aware refinement mechanism that selectively enhances poorly restored regions, improving overall fidelity and perceptual quality without compromising well-recovered areas.
\end{itemize}

\section{Related Works}
\label{sec:relatedWorks}

\begin{figure*}[t]
\centering
\includegraphics[width=\textwidth]{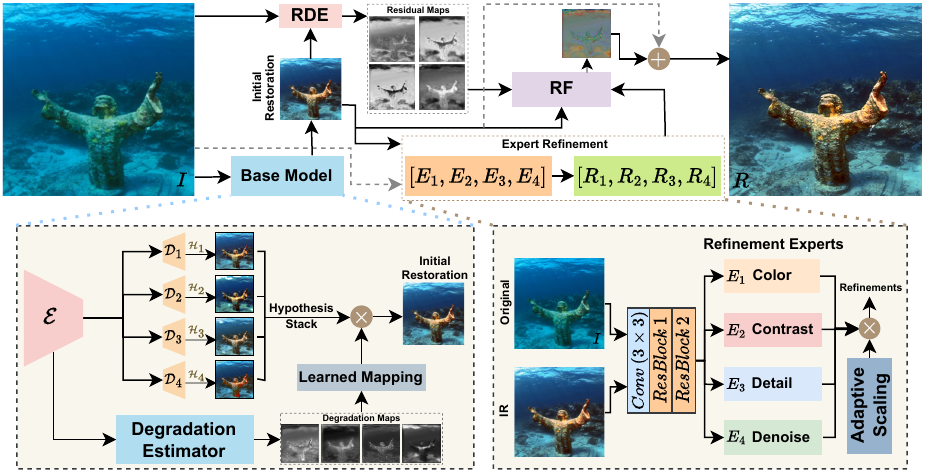}
\caption{\textbf{Architecture of the TIDE framework.} The top shows the overall flow: the input image is first processed by the base model to generate an initial restoration, which along with the original image feeds into the residual degradation estimator (RDE) and refinement fusion (RF) to produce the final output. The bottom left shows the base model details, where a feature extractor $\mathcal{E}$ feeds specialized decoders $\mathcal{D}_1$-$\mathcal{D}_4$ that generate restoration hypotheses $\mathcal{H}_1$-$\mathcal{H}_4$, while also estimating degradation maps. These hypotheses are combined through a learned mapping function for initial restoration. The bottom right illustrates the refinement stage, where the original and initially restored images are processed by shared convolutional processing before being fed to specialized refinement experts $E_1$-$E_4$ that generate targeted corrections for color, contrast, detail, and noise, which are then adaptively scaled and combined.}
\label{fig:architecture}
\end{figure*}

\paragraph{Underwater Image Restoration.} 
UIR typically relied on traditional image processing techniques such as histogram equalization, white balance correction, and physics-based dehazing methods~\cite{turn0search0, turn0search6, turn0search5}. While these methods improve visual contrast and partially correct color distortions, they often struggle to generalize across diverse underwater scenes and miss fine image details. Deep learning 
 solutions have achieved significant improvements~\cite{qiao2023underwater,li2019underwater,islam2020fast,li2021underwater,wu2021two,tolie2024dicam,tao2024multi,lai2023two}. GANs, in particular, enable realistic dataset synthesis and unpaired image enhancement~\cite{guo2019underwater,goodfellow2014generative,liu2019underwater}. Ucolor addresses color distortion and poor contrast using a multi-color space encoder with attention~\cite{li2021underwater}, FUSION leverages spatial and frequency domain information for effective fusion and color recalibration~\cite{fusion}, and FUnIE-GAN and Water-Net restore color fidelity and perceptual realism via adversarial optimization and confidence-guided fusion~\cite{turn0search7, turn0search8}.

\vspace{-3mm}

\paragraph{Multi-Hypothesis and Progressive Image Recovery.} 

Progressive architectures and multi-hypothesis strategies, such as recursive residual networks and staged refinement modules, have proven effective for complex image restoration by simplifying optimization and improving accuracy~\cite{4,5}. Recent work extends this by explicitly modeling degradation or domain characteristics, \textit{e.g.,} ingredient-aware reformulations~\cite{6}, severity-aware weather removal~\cite{7}, and multi-weather restoration via domain translation~\cite{3}. In the super-resolution domain, multi-hypothesis techniques have been used to fuse spatially offset patches and improve restoration of compressed screen content~\cite{1}. Other approaches combine progressive designs with semantic priors or reference images to tackle specific tasks, such as face restoration through semantic-aware style transfer~\cite{9}, reference-guided multi-degradation refinement~\cite{11}, and staged fusion for HDR reconstruction with attention~\cite{10}. Task-specific systems like PRISM progressively reconstruct scene graph image edits~\cite{2}, while lensless image restoration frameworks integrate multi-stage supervision with GAN refinement~\cite{8}.

\section{Method}
\label{sec:method}

\subsection{Overview}

We approach UIR via inverse degradation estimation and prior disentanglement. Instead of directly mapping degraded images $\mathbf{I}$ to clean ones, we explicitly model degradations and apply targeted restoration. The process is formulated as $\hat{\mathbf{J}} = \mathcal{R}(\mathbf{I}, \mathcal{D}(\mathbf{I}))$, where $\hat{\mathbf{J}}$ is the restored image, $\mathcal{R}$ is the restoration function, and $\mathcal{D}$ estimates degradation maps $\mathbf{M} = \{M_1, \dots, M_K\}$ for $K$ degradation types.  

Different degradations require specialized treatment. Through prior disentanglement, we generate multiple hypotheses $\mathbf{H} = \{H_1, \dots, H_K\}$, each addressing a specific degradation via $H_i = \mathcal{F}_i(\mathbf{I})$, where $\mathcal{F}_i$ is the corresponding specialized function. TIDE implements this in two stages: first, degradation-guided multi-hypothesis restoration combines specialized hypotheses (\S~\ref{subsec:invdeg}); second, a refinement stage corrects residual degradations (\S~\ref{subsec:twostage}). Fig.~\ref{fig:architecture} illustrates the architecture with its targeted components and progressive refinement.

\subsection{Inverse Degradation Mapping with Specialized Prior Decomposition}
\label{subsec:invdeg}

The first stage of TIDE performs inverse degradation mapping to identify the spatial distribution and severity of different degradation types, followed by specialized prior decomposition to generate targeted restoration hypotheses. 

\vspace{-3mm}

\paragraph{Degradation Estimation.}

To restore underwater images effectively, we first identify the degradations affecting each spatial location. We design a lightweight encoder-decoder network $\mathcal{D}$ that maps an input image $\mathbf{I} \in \mathbb{R}^{3 \times H \times W}$ to $K$ degradation maps $\mathbf{M} = \mathcal{D}(\mathbf{I}) \in \mathbb{R}^{K \times H \times W}$, where each element $M_k(x,y) \in [0,1]$ indicates the severity of degradation type $k$ at spatial location $(x,y)$. The network uses a multi-scale architecture with skip connections to capture both local and global patterns. It consists of an initial feature extraction layer $f_0$, a series of downsampling layers $\{f_d^i\}_{i=1}^N$, a global context module $f_g$, and upsampling layers $\{f_u^i\}_{i=1}^N$ with skip connections. The global context module aggregates spatial information via global average pooling followed by channel-wise modulation. The output degradation maps are generated through a sigmoid activation, $\mathbf{M} = \sigma\big(W_o f_u^1(f_u^2(\cdots f_u^N(\mathbf{g}, \mathbf{z}_{N-1}) \cdots), \mathbf{z}_0)\big)$, where $W_o$ is the output projection and $f_u^i(\mathbf{a}, \mathbf{b})$ denotes the $i$-th upsampling layer with skip connection from feature map $\mathbf{b}$.

\vspace{-3mm}

\paragraph{Inverse Degradation Mapping with Specialized Prior Decomposition.}

We decompose underwater image degradation into four fundamental types - color distortion, contrast reduction, detail loss, and noise - motivated by the underlying imaging physics~\cite{uwprop1,uwprop2,uwprop3}. Color distortion arises from wavelength-dependent attenuation, where longer wavelengths (red) diminish faster than shorter ones (blue/green)~\cite{uwprop4}. The reduction in contrast occurs due to the scattering of light by suspended particles, which produces haze and reduces the dynamic range~\cite{uwprop5}. Detail loss results from absorption and forward scattering that blur fine structures~\cite{uwprop6}, while noise originates from sensor limitations in low-light conditions and scattering effects such as marine snow~\cite{uwprop7}. While other degradation types such as extreme low-light or turbidity-related scatter could be considered, our experiments show these are effectively captured by the existing experts: low-light effects manifest primarily as contrast reduction and increased noise, and turbidity-related scatter is captured by the noise decoder. Our ablations (Table~\ref{tab:ablation}) justify that these four types of degradation form a minimal but sufficient set for effective restoration, balancing restoration quality with computational efficiency while avoiding unnecessary complexity.

\vspace{-3mm}

\paragraph{Specialized Restoration Decoders.}  

We design $K=4$ specialized decoders, each targeting one degradation type: color distortion, contrast reduction, detail loss, and noise. All decoders share a common feature extraction backbone $\mathcal{E}$, which generates hierarchical representations $\mathbf{F} = \{\mathbf{f}_0, \mathbf{f}_1, \ldots, \mathbf{f}_N\}$ from the input image $\mathbf{I}$, i.e., $\mathbf{F} = \mathcal{E}(\mathbf{I})$. Each specialized decoder produces a restoration hypothesis $H_k = \mathcal{F}_k(\mathbf{F})$ for $k \in \{1,2,3,4\}$, with architectural inductive biases that guide them toward their target degradation.

The color restoration decoder $\mathcal{F}_1$ uses channel-wise attention ($\mathbb{R}^C \rightarrow \mathbb{R}^{C/r} \rightarrow \mathbb{R}^C$) to correct wavelength-dependent attenuation. The contrast enhancement decoder $\mathcal{F}_2$ employs residual blocks $\phi: \mathbb{R}^{C \times H \times W} \rightarrow \mathbb{R}^{C \times H \times W}$ to address scattering-induced contrast loss. The detail recovery decoder $\mathcal{F}_3$ implements cascaded residual blocks with cumulative enhancement $\mathcal{E}_d(\mathbf{x}) = \mathbf{x} + \sum_{i=1}^{L} \mathcal{R}_i(\mathbf{x})$ to recover fine structures lost through medium absorption. The denoising decoder $\mathcal{F}_4$ applies group convolutions with $g = \min(\lfloor C/8 \rfloor, C)$ partitions to remove noise from suspended particles.

\begin{figure}[t]
    \centering

    \makebox[0.3\linewidth]{Degraded}
    \hfill
    \makebox[0.3\linewidth]{Initial Restoration}
    \hfill
    \makebox[0.3\linewidth]{Refinement}

    \vspace{0.5em}

    \begin{subfigure}{0.3\linewidth}
        \includegraphics[width=\linewidth]{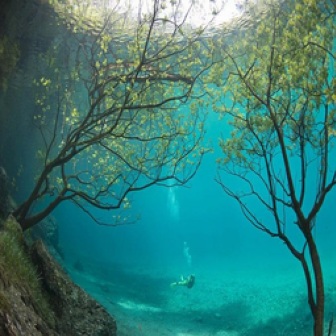}
    \end{subfigure}
    \hfill
    \begin{subfigure}{0.3\linewidth}
        \includegraphics[width=\linewidth]{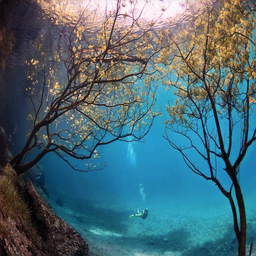}
    \end{subfigure}
    \hfill
    \begin{subfigure}{0.3\linewidth}
        \includegraphics[width=\linewidth]{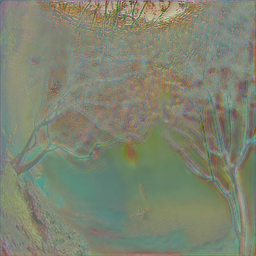}
    \end{subfigure}

    \vspace{0.5em}

    \begin{subfigure}{0.3\linewidth}
        \includegraphics[width=\linewidth]{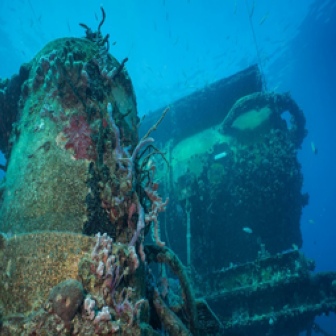}
    \end{subfigure}
    \hfill
    \begin{subfigure}{0.3\linewidth}
        \includegraphics[width=\linewidth]{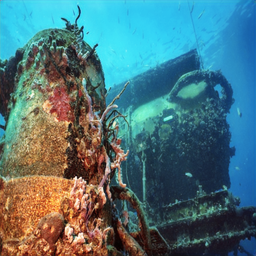}
    \end{subfigure}
    \hfill
    \begin{subfigure}{0.3\linewidth}
        \includegraphics[width=\linewidth]{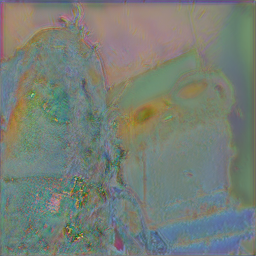}
    \end{subfigure}

    \vspace{0.5em}

    \begin{subfigure}{0.3\linewidth}
        \includegraphics[width=\linewidth]{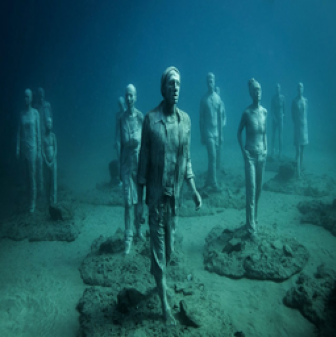}
    \end{subfigure}
    \hfill
    \begin{subfigure}{0.3\linewidth}
        \includegraphics[width=\linewidth]{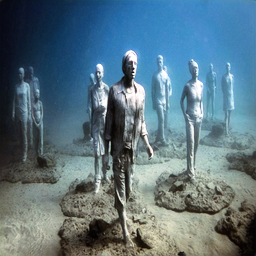}
    \end{subfigure}
    \hfill
    \begin{subfigure}{0.3\linewidth}
        \includegraphics[width=\linewidth]{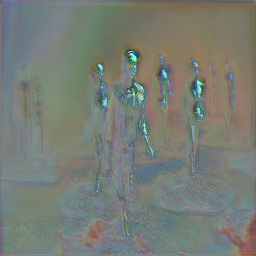}
    \end{subfigure}

    \caption{Visual comparison of input degradations, initial restoration, and expert refinement stages, illustrating TIDE’s progressive restoration process.}
    \label{fig:qualitative_results}
\end{figure}

This specialization is reinforced during training via two losses. The diversity loss $\mathcal{L}_{div}(\mathbf{H}) = \frac{1}{N_{pairs}} \sum_{i=1}^{K} \sum_{j=i+1}^{K} \text{cos\_sim}(H_i, H_j)$ regularizes similarity between hypotheses, preventing redundant representations, and the degradation consistency loss $\mathcal{L}_{dc}(\mathbf{M}, \mathbf{I}, \mathbf{J}) = \mathcal{L}_{MSE}(\sum_{k=1}^{K} M_k, \text{norm}(|\mathbf{I} - \mathbf{J}|))$ aligns estimated degradation maps with actual degradation patterns. Together, these losses allow each decoder to develop optimal strategies for its target degradation without requiring explicit labels, enabling the system to handle complex, co-occurring degradations effectively.


\vspace{-3mm}

\paragraph{Adaptive Fusion Mechanism.}

Given the $K$ specialized restoration hypotheses $\{H_1, H_2, \ldots, H_K\}$ and the degradation maps $\mathbf{M}$, we perform an adaptive fusion to combine them into the final restored image. The fusion process generates pixel-wise weights $\mathbf{W} \in \mathbb{R}^{K \times H \times W}$ that determine the contribution of each hypothesis at each spatial location:

\begin{equation}
\hat{\mathbf{J}}_1 = \sum_{k=1}^{K} W_k \odot H_k
\end{equation}

\noindent where $W_k$ is the weight map for hypothesis $H_k$, and $\odot$ represents element-wise multiplication. We implement three fusion strategies, with the learned fusion being our primary approach. The learned fusion mechanism transforms degradation maps into fusion weights through a non-linear mapping. This enables the model to select the appropriate restoration strategy for each image region based on its specific degradation characteristics.

\subsection{Two-Stage Degradation Compensation with Expert-Guided Correction}
\label{subsec:twostage}

Our second stage implements a progressive refinement approach that explicitly identifies and addresses these remaining artifacts through differential degradation analysis and expert-guided correction.

\vspace{-3mm}

\paragraph{Differential Degradation Analysis.}

Residual degradations are identified by estimating discrepancies between the initial restoration $\hat{\mathbf{J}}_1$ and the degraded input $\mathbf{I}$. We implement this through a differential degradation analysis function $\mathcal{D}_2$, which predicts spatial residuals: $\mathbf{M}_r = \mathcal{D}_2(\mathbf{I}, \hat{\mathbf{J}}_1)$, where $\mathbf{M}_r \in \mathbb{R}^{K \times H \times W}$ represents the residual degradation maps for the $K$ degradation types. First, it computes an absolute difference map $\mathbf{D} = |\mathbf{I} - \hat{\mathbf{J}}_1|$ to localize regions with remaining degradations. Since not all discrepancies indicate incomplete restoration, we process the concatenated input $[\mathbf{I}, \hat{\mathbf{J}}_1] \in \mathbb{R}^{6 \times H \times W}$ through an attention network guided by $\mathbf{D}$:

\begin{equation}
\mathbf{M}_r = \sigma(f_r([\mathbf{I}, \hat{\mathbf{J}}_1]) \cdot (1 + \alpha \cdot f_d(\mathbf{D})))
\end{equation}

\noindent where $f_r$ is the residual estimation network, $f_d$ is the difference enhancement network, $\alpha$ is a learnable scaling factor, and $\sigma$ is the sigmoid function. This allows the network to focus on regions with significant residual degradations while considering contextual information from both the input and initial restoration.

\vspace{-3mm}

\paragraph{Expert-Guided Correction Generation.}

Based on the residual degradation maps, we generate targeted corrections for each degradation type using specialized refinement experts. Each expert \(\mathcal{E}_k\) takes the original image and initial restoration as inputs and produces a correction term \(C_k = \mathcal{E}_k(\mathbf{I}, \hat{\mathbf{J}}_1)\), \(k \in \{1,2,3,4\}\), specific to its degradation type. We design four refinement experts corresponding to the same degradation types addressed in the first stage. Each expert implements specialized processing tailored to its target degradation, generating a correction term rather than a complete restoration. To prevent excessive corrections that might introduce artifacts, each expert includes an adaptive scaling mechanism $C_k = \tanh(f_k(\mathbf{I}, \hat{\mathbf{J}}_1)) \cdot \sigma(s_k)$, where $f_k$ denotes the expert's processing function, $\tanh$ constrains the correction range to $[-1, 1]$, and $s_k$ is a learnable scaling parameter controlling the correction magnitude.

\vspace{-3mm}

\paragraph{Safety-Gated Fusion.}

The correction terms from individual experts are then carefully combined to avoid over-correction. We implement a safety-gated fusion mechanism that considers both the residual degradation maps and the quality of the initial restoration:

\begin{equation}
\mathbf{C} = \left(\sum_{k=1}^{K} (M_r)_k \odot C_k\right) \cdot G(\hat{\mathbf{J}}_1) \cdot \sigma(s_g)
\end{equation}

\noindent where $(M_r)_k$ is the residual map for degradation type $k$, $G$ is a safety gate function that limits corrections in well-restored regions, and $s_g$ is a global scaling factor. The safety gate $G$ examines the initial restoration $\hat{\mathbf{J}}_1$ and produces a spatial mask that modulates the correction intensity, computed as $G(\hat{\mathbf{J}}_1) = \sigma(f_g(\hat{\mathbf{J}}_1))$, where $f_g$ consists of convolutional layers that identify regions requiring minimal correction. The final restoration is then obtained by adding the fused correction map $\mathbf{C}$ to the initial restoration as $\hat{\mathbf{J}} = \hat{\mathbf{J}}_1 + \mathbf{C}$, followed by a clamping operation to ensure that the pixel values remain within the valid range $[0, 1]$.

\vspace{-3mm}

\paragraph{Optimization.}

We train our two-stage framework with a structured two-phase optimization. In stage 1, we optimize only the base model with reconstruction, diversity, and degradation-consistency losses: 
\(\mathcal{L}_1 = \lambda_{\ell_1}\mathcal{L}_{\ell_1} + \lambda_{ssim}\mathcal{L}_{ssim} + \lambda_p\mathcal{L}_p + \lambda_{div}\mathcal{L}_{div} + \lambda_{dc}\mathcal{L}_{dc}\), 
where \(\mathcal{L}_{\ell_1}\), \(\mathcal{L}_{ssim}\), \(\mathcal{L}_p\) are L1, SSIM, and perceptual reconstruction losses, \(\mathcal{L}_{div}\) encourages diverse hypotheses, and \(\mathcal{L}_{dc}\) enforces consistency between predicted and actual degradations. In stage 2, the base model is frozen and the refinement components are trained with a progressive improvement loss: 
\(\mathcal{L}_2 = \mathcal{L}_{recon} + \lambda_{mag}\|\mathbf{C}\|_1 + \lambda_{imp}\mathcal{L}_{imp}\), 
where \(\mathcal{L}_{recon}\) combines L1, SSIM, and perceptual losses, \(\|\mathbf{C}\|_1\) penalizes large corrections, and \(\mathcal{L}_{imp} = \max\big(0, \|\hat{\mathbf{J}} - \mathbf{J}\|_1 - \|\hat{\mathbf{J}}_1 - \mathbf{J}\|_1 + \epsilon\big)\) enforces explicit improvement over the base output.

\begin{figure*}[t]
    \centering
    
    \begin{adjustbox}{valign=c}
        \begin{sideways}
            \textbf{EUVP}
        \end{sideways}
    \end{adjustbox}%
    \begin{adjustbox}{valign=c}
        \includegraphics[width=0.98\linewidth]{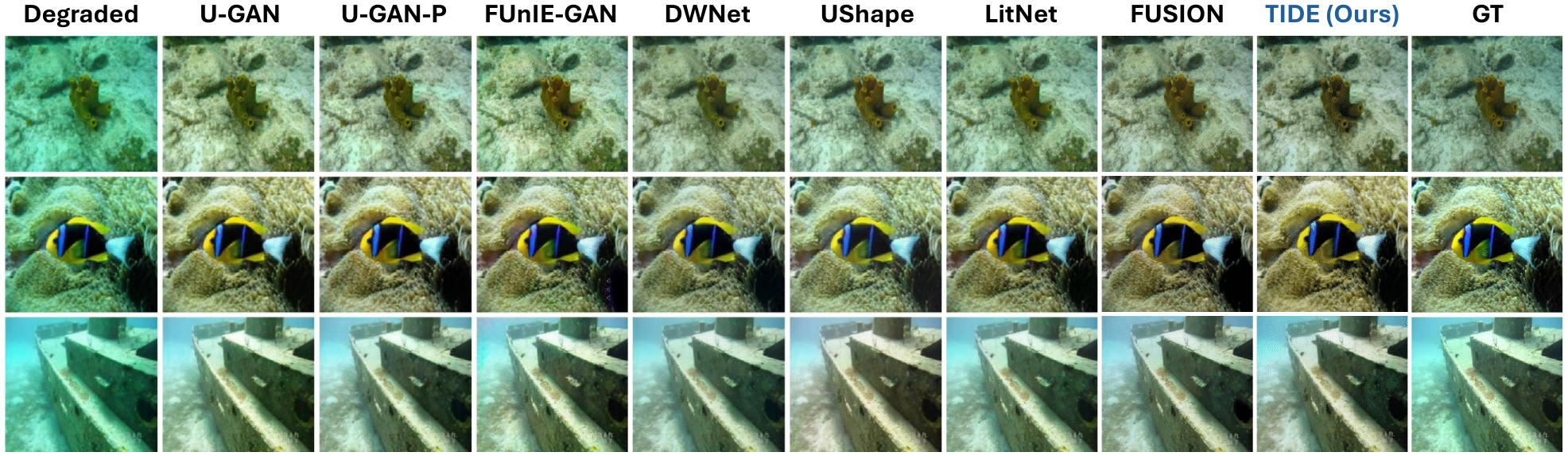}
    \end{adjustbox}
    
    \vspace{0.5em}
    
    \begin{adjustbox}{valign=c}
        \begin{sideways}
            \textbf{UIEB}
        \end{sideways}
    \end{adjustbox}%
    \begin{adjustbox}{valign=c}
        \includegraphics[width=0.98\linewidth]{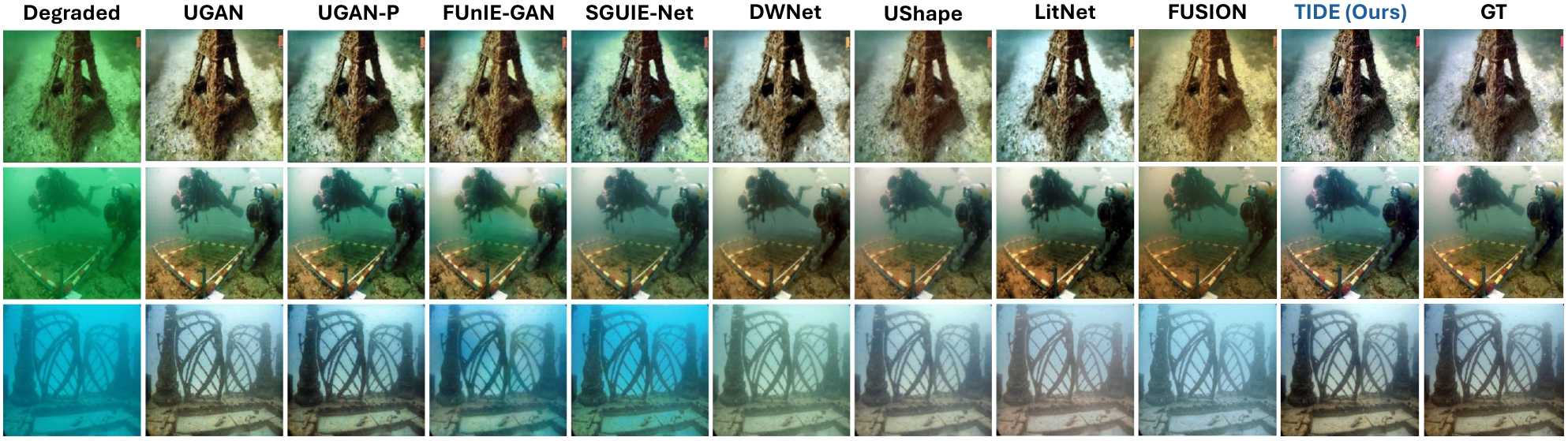}
    \end{adjustbox}
    
    \vspace{0.5em}
    
    \begin{adjustbox}{valign=c}
        \begin{sideways}
            \textbf{SUIM-E}
        \end{sideways}
    \end{adjustbox}%
    \begin{adjustbox}{valign=c}
        \includegraphics[width=0.98\linewidth]{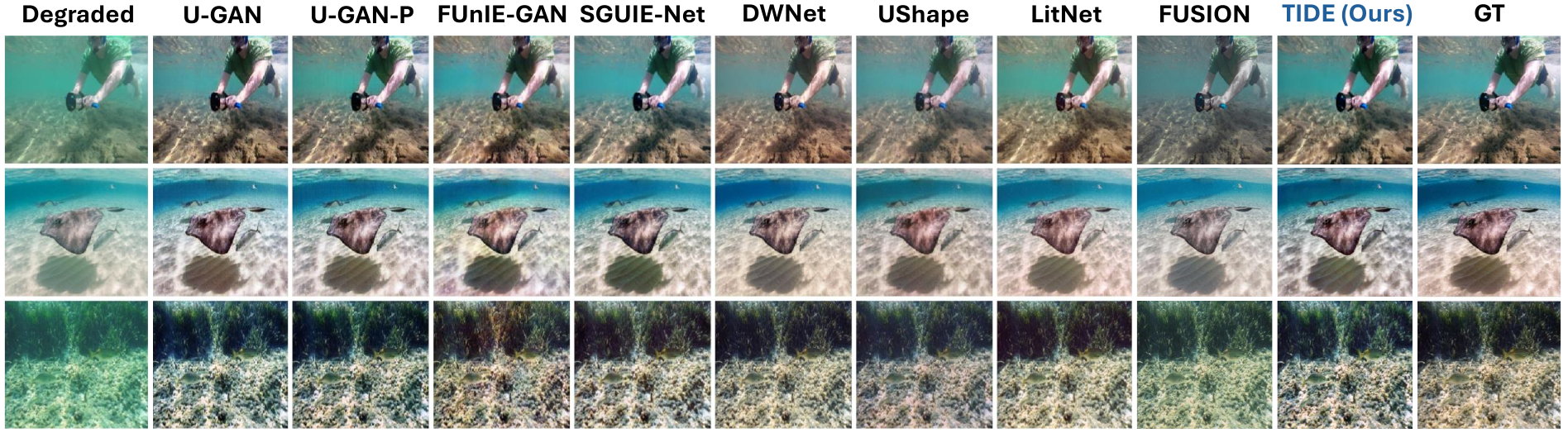}
    \end{adjustbox}

    \caption{Qualitative restoration results across EUVP, UIEB, and SUIM-E datasets.}
    \label{fig:qualitative_all}
\end{figure*}

\section{Experiments}
\label{sec:experiments}


\paragraph{Experimental Settings.}
We evaluate TIDE on three standard underwater image datasets: UIEB~\cite{li2019underwater}, EUVP~\cite{islam2020fast}, and SUIM-E~\cite{qi2022sguie}, with all images resized to $256\times256$ for training and testing. We compare against state-of-the-art UIR methods using both full-reference (PSNR, SSIM, LPIPS) and no-reference metrics (UIQM, UISM, BRISQUE), along with UICM and UIConM, which specifically measure color cast correction and local contrast enhancement (detailed experimental settings and metric definitions are provided in the supplementary).

\subsection{Comparison with State-of-the-Art}

Figure~\ref{fig:qualitative_all} and Table~\ref{tab:combined_uie_performance} show that TIDE achieves consistent improvements across diverse degradations. Existing methods often address specific degradations but fail on others: U-GAN variants restore color but leave haze, FUSION preserves contrast while losing fine details, and DWNet removes haze at the cost of color fidelity, reflecting the challenge of spatially varying, co-occurring degradations in underwater scenes. TIDE mitigates this issue by applying degradation-specific processing, yielding balanced restoration in all cases. In the EUVP ship scene (row 3), it recovers low contrast, color distortions, and fine details simultaneously without introducing artifacts. Similarly, in the grass and statue examples from UIEB, TIDE maintains fine textures and structural details while removing color casts, whereas other methods either over-saturate or retain residual haze. Quantitatively, TIDE achieves higher fidelity and perceptual quality, reaching 0.910 SSIM on UIEB (vs. 0.883 for FUSION) and 3.440 UIQM (vs. 3.432 for U-GAN). These results stem from inverse degradation mapping, which disentangles co-occurring distortions, and the refinement stage, which reduces residual artifacts.

\begin{table*}[t]
\centering
\caption{Performance comparison of UIE methods across EUVP, UIEB, and SUIM-E test sets. $\downarrow$ indicates lower is better. First, second, and third best performances are represented with \colorbox{red!20}{red}, \colorbox{blue!20}{blue}, and \colorbox{green!20}{green} background colors, respectively.}
\resizebox{\textwidth}{!}
{\renewcommand{\arraystretch}{1.1}
\begin{tabular}{l|cccccc|cccccc|cccccc}
\hline
& \multicolumn{6}{c|}{\textbf{EUVP}} & \multicolumn{6}{c|}{\textbf{UIEB}} & \multicolumn{6}{c}{\textbf{SUIM-E}} \\
\hline
Method & PSNR & SSIM & UIQM & LPIPS~↓ & UISM & BRISQUE~↓ & PSNR & SSIM & LPIPS~↓ & UIQM & UISM & BRISQUE~↓ & PSNR & SSIM & LPIPS~↓ & UIQM & UISM & BRISQUE~↓ \\
\hline
DWNet~\cite{sharma2023wavelength} & 28.654 & 0.835 & 3.042 & \cellcolor{green!20}{\textbf{0.173}} & \cellcolor{green!20}{\textbf{7.051}} & \cellcolor{green!20}{\textbf{30.856}} & 23.165 & 0.843 & 0.162 & 2.897 & 7.089 & 24.863 & 24.850 & \cellcolor{green!20}{\textbf{0.861}} & \cellcolor{green!20}{\textbf{0.133}} & 2.707 & 7.381 & 20.757 \\
UGAN~\cite{ugan} & 26.551 & 0.807 & 2.896 & 0.220 & 6.833 & 35.859 & 23.322 & 0.815 & 0.199 & \cellcolor{blue!20}{\textbf{3.432}} & 7.241 & 27.011 & 24.704 & 0.826 & 0.190 & 2.894 & 7.175 & 20.288 \\
UGAN-P~\cite{ugan} & 26.549 & 0.805 & 2.931 & 0.223 & 6.816 & 35.099 & 23.550 & 0.814 & 0.192 & 3.396 & 7.262 & 25.382 & 25.050 & 0.827 & 0.188 & 2.901 & 7.184 & \cellcolor{green!20}{\textbf{18.768}} \\
FUSION~\cite{fusion} & \cellcolor{green!20}{\textbf{28.671}} & \cellcolor{blue!20}{\textbf{0.862}} & \cellcolor{blue!20}{\textbf{3.220}} & 0.174 & \cellcolor{blue!20}{\textbf{7.048}} & \cellcolor{blue!20}{\textbf{29.547}} & \cellcolor{blue!20}{\textbf{23.717}} & \cellcolor{blue!20}{\textbf{0.883}} & \cellcolor{red!20}{\textbf{0.112}} & \cellcolor{green!20}{\textbf{3.414}} & \cellcolor{blue!20}{\textbf{7.429}} & \cellcolor{green!20}{\textbf{23.193}} & \cellcolor{green!20}{\textbf{25.989}} & 0.850 & \cellcolor{red!20}{\textbf{0.118}} & \cellcolor{blue!20}{\textbf{3.183}} & \cellcolor{blue!20}{\textbf{7.679}} & \cellcolor{blue!20}{\textbf{18.655}} \\
Lit-Net~\cite{ms2_litnet} & \cellcolor{red!20}{\textbf{29.477}} & \cellcolor{green!20}{\textbf{0.851}} & 3.027 & \cellcolor{blue!20}{\textbf{0.169}} & 7.011 & 32.109 & \cellcolor{green!20}{\textbf{23.603}} & \cellcolor{green!20}{\textbf{0.863}} & \cellcolor{green!20}{\textbf{0.130}} & 3.145 & \cellcolor{green!20}{\textbf{7.396}} & \cellcolor{blue!20}{\textbf{23.038}} & 25.117 & \cellcolor{blue!20}{\textbf{0.884}} & \cellcolor{red!20}{\textbf{0.118}} & \cellcolor{green!20}{\textbf{2.918}} & 7.368 & 19.602 \\
FUnIE-GAN~\cite{islam2020fast} & 26.220 & 0.792 & 2.971 & 0.212 & 6.892 & 30.912 & 21.043 & 0.785 & 0.173 & 3.250 & 7.202 & 24.522 & 23.590 & 0.825 & 0.189 & \cellcolor{green!20}{\textbf{2.918}} & 7.121 & 22.560 \\
Ushape~\cite{peng2023u} & 26.822 & 0.811 & \cellcolor{green!20}{\textbf{3.052}} & 0.187 & 6.843 & 35.648 & 21.084 & 0.744 & 0.220 & 3.161 & 7.183 & 24.128 & 22.647 & 0.783 & 0.213 & 2.873 & 7.061 & 22.876 \\
SGUIE-Net~\cite{qi2022sguie} & - & - & - & - & - & - & 23.496 & 0.853 & 0.136 & 3.004 & 7.362 & 24.607 & \cellcolor{red!20}{\textbf{25.987}} & 0.857 & 0.153 & 2.637 & 7.090 & 25.927 \\
ULAP~\cite{song2018rapid} & - & - & - & - & - & - & 19.863 & 0.724 & 0.256 & 2.328 & 7.362 & 25.113 & 19.148 & 0.744 & 0.231 & 2.115 & \cellcolor{green!20}{\textbf{7.475}} & 21.250 \\
TIDE (Ours) & \cellcolor{blue!20}{\textbf{29.469}} & \cellcolor{red!20}{\textbf{0.906}} & \cellcolor{red!20}{\textbf{3.281}} & \cellcolor{red!20}{\textbf{0.159}} & \cellcolor{red!20}{\textbf{7.066}} & \cellcolor{red!20}{\textbf{29.545}} & \cellcolor{red!20}{\textbf{23.753}} & \cellcolor{red!20}{\textbf{0.910}} & \cellcolor{blue!20}{\textbf{0.115}} & \cellcolor{red!20}{\textbf{3.440}} & \cellcolor{red!20}{\textbf{7.439}} & \cellcolor{red!20}{\textbf{22.991}} & \cellcolor{blue!20}{\textbf{25.987}} & \cellcolor{red!20}{\textbf{0.906}} & \cellcolor{blue!20}{\textbf{0.119}} & \cellcolor{red!20}{\textbf{3.195}} & \cellcolor{red!20}{\textbf{7.685}} & \cellcolor{red!20}{\textbf{18.605}} \\
\hline
\end{tabular}
}
\label{tab:combined_uie_performance}
\end{table*}

\begin{figure*}[t]
\centering
\includegraphics[width=\textwidth]{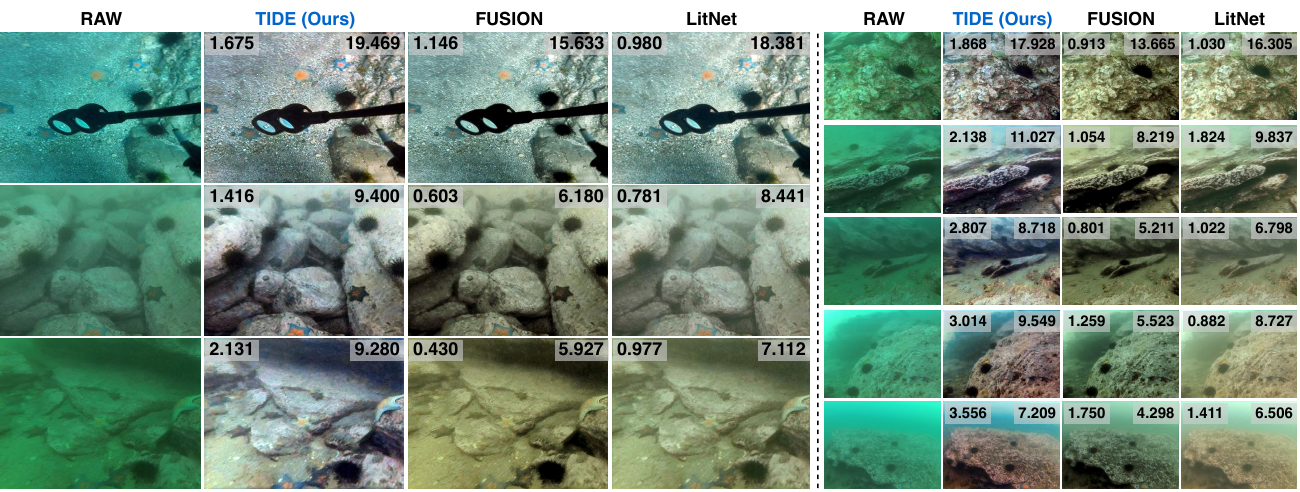}
\caption{Qualitative comparison on naturally turbid underwater images from the RUIE dataset. The left group corresponds to the UCCS dataset, while the right group corresponds to the UIQS dataset. For each result, the two metrics displayed in the top corners are UICM (left, measuring color correction) and UIConM (right, measuring contrast enhancement). Higher values indicate better performance for both metrics.}
\label{fig:turbid_results}
\end{figure*}

To evaluate TIDE under real-world conditions beyond clear-water datasets, we test it on the RUIE dataset~\cite{turbidDataset} using weights trained on UIEB. We focus on the UIQS and UCCS subsets, which capture naturally turbid scenes with strong color casts. UIEB was chosen for training as it provides sufficient paired data to learn robust enhancement mappings while remaining limited enough to test generalization. For comparison, we include FUSION~\cite{fusion} and LitNet~\cite{ms2_litnet}, the two strongest baselines from Table~\ref{tab:combined_uie_performance} to assess both relative performance and generalization. As shown in Fig.~\ref{fig:turbid_results}, TIDE consistently achieves higher UICM (2.13 vs. 0.43 and 0.98) and UIConM (9.28 vs. 5.93 and 7.11) scores. Qualitatively, it better handles heavily distorted regions: in row 1, TIDE restores natural rock textures while correcting the green cast, whereas FUSION introduces yellowish artifacts and LitNet leaves residual haze; in rows 4–5, it yields finer details of marine life with more natural coloration.

\subsection{Hardware Efficiency Analysis}

\begin{table*}
\centering
\caption{Ablation study of TIDE components. Removing specialized decoders, loss terms, or the refinement stage reduces performance across multiple datasets. The metrics displayed are an \textbf{average} over all the images in the test set.}
\resizebox{\textwidth}{!}{
{\renewcommand{\arraystretch}{1.1}
\begin{tabular}{c|cccc|ccc|ccc|ccc|cc|cc}
\toprule
\multicolumn{5}{c|}{{\textbf{Config}}} & \multicolumn{3}{c|}{\textbf{EUVP}} & \multicolumn{3}{c|}{\textbf{UIEB}} & \multicolumn{3}{c|}{\textbf{SUIM-E}} & \multicolumn{2}{c|}{\textbf{UIQS}} & \multicolumn{2}{c}{\textbf{UCCS}} \\
\cmidrule{1-18}
\multicolumn{1}{c|}{} & color & contrast & detail & denoise & \textbf{LPIPS~$\downarrow$} & \textbf{PSNR} & \textbf{SSIM} & \textbf{LPIPS~$\downarrow$} & \textbf{PSNR} & \textbf{SSIM} & \textbf{LPIPS~$\downarrow$} & \textbf{PSNR} & \textbf{SSIM} & \textbf{UICM} & \textbf{UIConM} & \textbf{UICM} & \textbf{UIConM} \\
\cmidrule{2-18}
\multirow{8}{*}{\rotatebox[origin=c]{90}{\textbf{\hspace{12mm}Decoder Types}}} 
& \ding{51} & \ding{55} & \ding{55} & \ding{55} & 0.216 & 22.391 & 0.867 & 0.158 & 22.809 & 0.890 & 0.158 & 22.517 & 0.882 & 13.451 & 0.847 & 13.563 & 0.862 \\
& \ding{51} & \ding{51} & \ding{55} & \ding{55} & 0.209 & 22.623 & 0.873 & 0.205 & 23.333 & 0.901 & 0.204 & 23.216 & 0.900 & 13.642 & 0.894 & 13.731 & 0.908 \\
& \ding{51} & \ding{55} & \ding{51} & \ding{55} & 0.210 & 22.565 & 0.872 & 0.207 & 23.239 & 0.899 & 0.206 & 23.096 & 0.898 & 13.598 & 0.885 & 13.694 & 0.901 \\
& \ding{51} & \ding{55} & \ding{55} & \ding{51} & 0.207 & 22.713 & 0.876 & 0.203 & 23.210 & 0.899 & 0.207 & 22.853 & 0.892 & 13.683 & 0.909 & 13.768 & 0.923 \\
& \ding{51} & \ding{51} & \ding{51} & \ding{55} & 0.192 & 23.252 & 0.889 & 0.203 & 23.494 & 0.904 & 0.201 & 23.365 & 0.903 & 14.087 & 1.012 & 14.195 & 1.031 \\
& \ding{51} & \ding{51} & \ding{55} & \ding{51} & 0.195 & 23.194 & 0.888 & 0.201 & 23.488 & 0.902 & 0.202 & 23.202 & 0.900 & 14.024 & 0.991 & 14.138 & 1.013 \\
& \ding{51} & \ding{55} & \ding{51} & \ding{51} & 0.190 & 23.331 & 0.891 & 0.200 & 23.484 & 0.903 & 0.205 & 23.101 & 0.900 & 14.145 & 1.024 & 14.253 & 1.045 \\
\midrule
\multirow{2}{*}{\rotatebox[origin=c]{90}{\textbf{Loss}}} 
& \multicolumn{4}{c|}{No Degradation Consistency} & 0.354 & 19.841 & 0.792 & 0.228 & 21.095 & 0.802 & 0.252 & 21.626 & 0.792 & 13.205 & 0.813 & 13.458 & 0.826 \\
& \multicolumn{4}{c|}{No Diversity} & 0.229 & 21.231 & 0.845 & 0.170 & 22.701 & 0.886 & 0.189 & 21.853 & 0.861 & 13.894 & 0.962 & 14.127 & 0.981 \\
\midrule
\multirow{4}{*}{\rotatebox[origin=c]{90}{\textbf{Supervision}}} 
& \multicolumn{4}{c|}{Direct Fusion} & 0.200 & 22.973 & 0.882 & 0.202 & 23.461 & 0.903 & 0.221 & 23.042 & 0.897 & 14.162 & 1.024 & 14.297 & 1.047 \\
& \multicolumn{4}{c|}{No Degradation Maps} & 0.223 & 22.022 & 0.859 & 0.295 & 21.435 & 0.844 & 0.255 & 23.259 & 0.902 & 13.733 & 0.881 & 13.946 & 0.902 \\
& \multicolumn{4}{c|}{Single Hypothesis} & 0.202 & 22.868 & 0.880 & 0.210 & 22.740 & 0.888 & 0.210 & 22.494 & 0.884 & 13.985 & 0.953 & 14.104 & 0.968 \\
& \multicolumn{4}{c|}{No Refinement} & 0.188 & 23.414 & 0.892 & 0.202 & 23.547 & 0.904 & 0.202 & 23.417 & 0.905 & 14.294 & 1.065 & 14.385 & 1.089 \\
\midrule
\rowcolor[gray]{0.8}
\multicolumn{5}{c|}{\textbf{Full TIDE}} & \textbf{0.159} & \textbf{29.469} & \textbf{0.906} & \textbf{0.115} & \textbf{23.753} & \textbf{0.910} & \textbf{0.119} & \textbf{25.987} & \textbf{0.906} & \textbf{14.647} & \textbf{1.134} & \textbf{14.729} & \textbf{1.107} \\
\bottomrule
\end{tabular}}
}
\label{tab:ablation}
\end{table*}

\begin{figure}[t]
\centering
\includegraphics[width=\columnwidth]{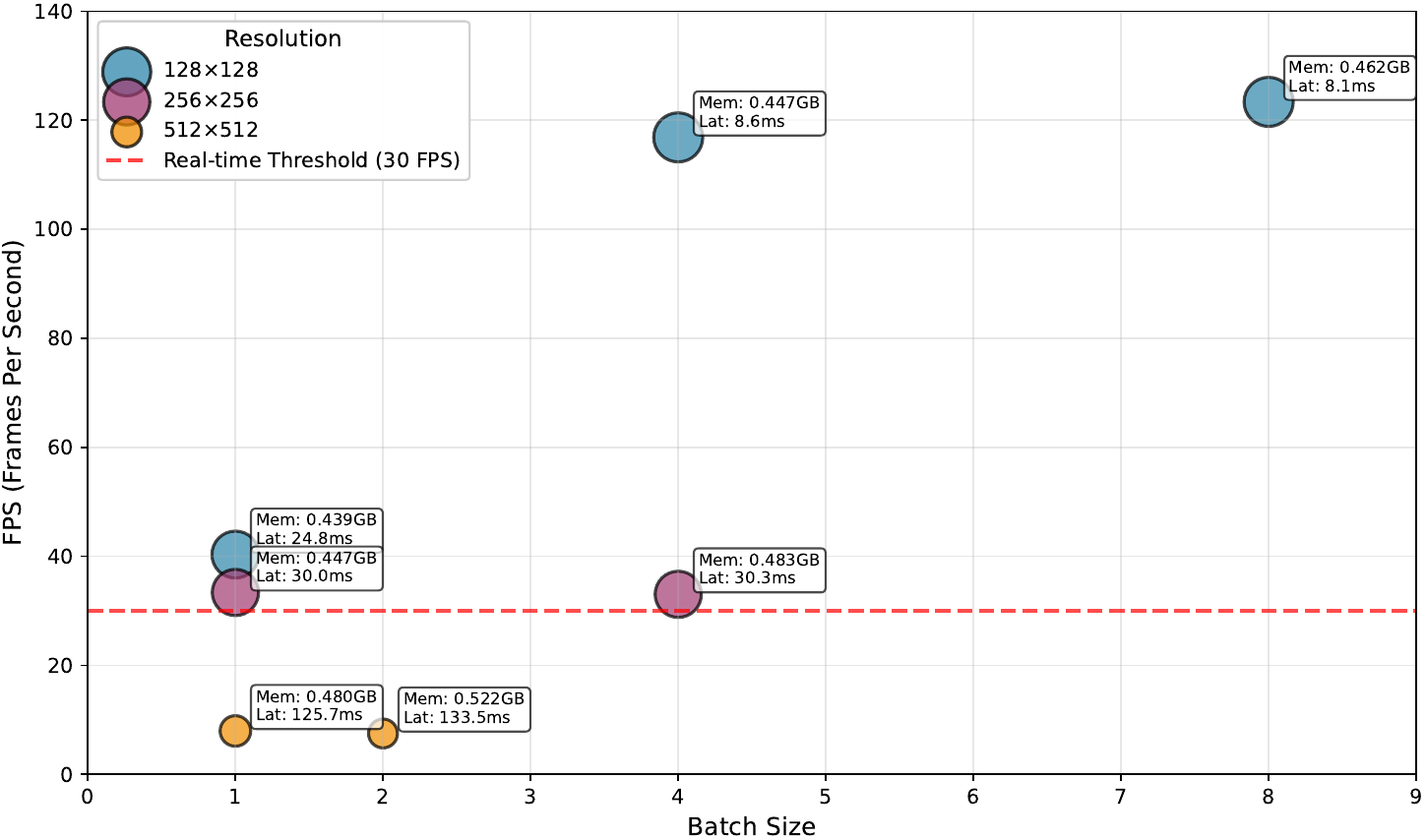}
\caption{Performance analysis of TIDE across different resolutions and batch sizes on RTX 4070 Ti SUPER. Bubble size indicates processing speed (inverse latency), while position represents the throughput-memory tradeoff. The horizontal red line marks the 30 FPS real-time processing threshold. Even at $256\times256$ resolution, TIDE achieves real-time performance (33.7 FPS) with minimal memory usage (0.44 GB), while smaller resolutions enable substantial throughput improvements with negligible overhead.}
\label{fig:performance_analysis}
\end{figure}

TIDE achieves remarkable inference efficiency through its architectural design, even with its substantial model size of 115.8M parameters. We evaluate performance on an RTX 4070 Ti SUPER to demonstrate computational efficiency on high-end consumer hardware, representing the upper bound of deployment performance potential. Figure~\ref{fig:performance_analysis} demonstrates that our model achieves real-time performance ($>$30 FPS) at $256\times256$ resolution with a single batch, and significantly higher throughput (123.4 FPS) at $128\times128$ resolution with batch size 8, while maintaining minimal memory footprint (0.46 GB). This efficiency arises from three design choices: (1) decoupled specialized decoders that process shared features in parallel rather than sequentially; (2) a refinement stage that introduces only 1.1\% overhead (1.25M params) by generating sparse corrections instead of full restorations; and (3) degradation-guided fusion that selectively applies computation to degraded regions, avoiding redundant processing in well-preserved areas. This enables TIDE to scale efficiently with batch size: throughput increases from 40.3 FPS to 123.4 FPS when moving from batch size 1 to 8 at $128\times128$ resolution with only a marginal 0.02 GB memory increase, while at $256\times256$ resolution, the model maintains consistent real-time performance ($\sim$33 FPS) when scaling from batch size 1 to 4 with minimal memory overhead (0.04 GB~$\uparrow$), making it well-suited for both real-time applications and high-throughput batch processing scenarios on modern consumer hardware.

\subsection{Ablation Study}

We conduct a comprehensive ablation study to assess the contribution of each component (Table~\ref{tab:ablation}). For the decoders, a color-only variant reaches 22.39 dB PSNR on EUVP but struggles with diverse degradations. Adding additional decoders consistently improves performance, with certain combinations showing clear complementarity: color–detail–denoise achieves 23.33 dB PSNR, slightly higher than color–contrast–detail at 23.25 dB. Similar behavior is observed on turbid datasets, where color–detail–denoise obtains 14.15 and 14.25 UICM on UIQS and UCCS, compared to 13.45 and 13.56 for color-only, indicating that dedicated processing paths are more effective than a single-purpose design. Loss function ablations show that removing degradation consistency or diversity losses substantially reduces performance across all datasets (\textit{e.g.}, LPIPS rises from 0.159 to 0.354 on EUVP without consistency), with particularly large drops on turbid water (UICM decreases from 14.65 to 13.21 on UIQS). Architectural ablations show similar trends: direct fusion in place of adaptive fusion, omitting degradation maps, or using a single hypothesis each reduce performance. Removing the refinement stage results in the most pronounced degradation, with EUVP PSNR dropping by 6.06 dB and UIQS UICM decreasing from 14.65 to 14.29. 

\vspace{1mm}

\textbf{Limitations.} 
While TIDE demonstrates significant improvements over existing methods, some limitations remain. In scenes with extreme lighting variations (\textit{e.g.}, areas affected simultaneously by direct sunlight and deep shadows), degradation maps may not balance restoration perfectly, potentially causing overprocessing in some regions while leaving subtle degradations in others. This stems from the inherent difficulty of disentangling heavily intertwined degradations rather than our four-category framework. Additionally, TIDE may struggle with uncommon underwater phenomena, such as complex caustics or unusual particulate scattering. Although evaluations on challenging real-world datasets show good generalization, performance could be further enhanced using physics-guided approaches that model these rare optical effects without extensive additional training data.

\section{Conclusion}

We introduced TIDE, a two-stage framework for UIR that addresses spatially varying degradations through inverse degradation mapping with specialized prior decomposition and progressive refinement. By explicitly modeling four fundamental degradation types, TIDE effectively handles the heterogeneous combination of color distortion, contrast reduction, detail loss, and noise in underwater imagery. Experiments across standard benchmarks and challenging turbid water scenarios demonstrate consistent performance improvements in both reference-based metrics and perceptual quality measures. We hope TIDE will inspire future research on more effective underwater image/video processing systems that adapt to diverse degradations, benefiting applications in underwater robotics and archaeological surveys.

{
    \small
    \bibliographystyle{ieeenat_fullname}
    \bibliography{main}
}

\clearpage
\setcounter{page}{1}
\maketitlesupplementary
\renewcommand{\thesection}{A\arabic{section}}
\setcounter{section}{0}

\section{Extended Methodology}
\label{sec:extended_method}

In this section, we provide detailed elaboration of our Two-stage Inverse Degradation Estimation (TIDE) framework. We describe the architectural components, fusion mechanisms, and optimization strategies that enable effective underwater image restoration through degradation-aware processing.

\subsection{Network Architecture and Component Details}
\label{subsec:architecture_details}

Our framework builds upon a multi-scale feature extraction backbone coupled with specialized processing modules. Here we provide a detailed examination of each component's internal structure and mathematical formulation.

The feature extraction module $\mathcal{E}$ employs a hierarchical encoder structure with $N$ downsampling operations, generating feature maps at progressively lower resolutions. For an input image $\mathbf{I} \in \mathbb{R}^{3 \times H \times W}$, the initial features are extracted via:

\begin{equation}
\mathbf{f}_0 = \phi(\mathbf{I}, \mathbf{W}_0, \mathbf{b}_0)
\end{equation}

\noindent where $\phi$ represents a convolution operation with normalization and non-linear activation, parameterized by weights $\mathbf{W}_0$ and bias $\mathbf{b}_0$. Subsequent feature maps are computed recursively as $\mathbf{f}_i = \psi(\mathbf{f}_{i-1}, \mathbf{W}_i, \mathbf{b}_i)$ for $i \in \{1, 2, ..., N\}$, where $\psi$ incorporates strided convolution for downsampling. At the bottleneck, we apply multiple residual blocks to enhance feature representation without altering spatial dimensions:

\begin{equation}
\mathbf{f}_N = \mathbf{f}_N + \sum_{j=1}^{R} \rho_j(\mathbf{f}_N)
\end{equation}

\noindent where $\rho_j$ represents the $j$-th residual block with $R=2$ in our implementation. The complete feature hierarchy $\mathbf{F} = \{\mathbf{f}_0, \mathbf{f}_1, ..., \mathbf{f}_N\}$ captures multi-scale information critical for addressing different degradation types.

The degradation estimation module $\mathcal{D}$ implements a lightweight encoder-decoder structure augmented with a global context module. The encoder follows a similar hierarchical design as $\mathcal{E}$ but with fewer channels, generating a compact feature representation $\mathbf{z}_N$. The global context module enhances this representation by modeling long-range dependencies:

\begin{equation}
\mathbf{g} = \mathbf{z}_N \odot \sigma(W_2(\delta(W_1(\text{GAP}(\mathbf{z}_N)))))
\end{equation}

\noindent where GAP denotes global average pooling, $W_1$ and $W_2$ are linear transformations, $\delta$ is ReLU activation, and $\sigma$ is sigmoid activation. This global context $\mathbf{g}$ is concatenated with $\mathbf{z}_N$ before progressive upsampling with skip connections from the encoder. The final degradation maps $\mathbf{M} \in \mathbb{R}^{K \times H \times W}$ are computed as:

\begin{equation}
\mathbf{M} = \sigma(W_o(\mathbf{u}_0))
\end{equation}

\noindent where $\mathbf{u}_0$ is the final upsampled feature map and $W_o$ is a $1 \times 1$ convolution projecting to $K$ channels, followed by sigmoid activation to ensure $M_k(x,y) \in [0,1]$.

\subsection{Degradation Types and Specialized Decoders}
\label{subsec:specialized_decoders}

We decompose underwater image degradation into four fundamental types based on the physical processes affecting underwater light propagation:

\begin{enumerate}
    \item \textbf{Color distortion}: Caused by wavelength-dependent attenuation where red light (longer wavelengths) is absorbed more rapidly than green and blue light (shorter wavelengths), resulting in the characteristic blue-green color cast of underwater images.
    
    \item \textbf{Contrast reduction}: Results from light scattering by suspended particles, creating haze and reducing the dynamic range of the image, particularly at greater depths or in turbid waters.
    
    \item \textbf{Detail loss}: Occurs due to medium absorption and forward scattering effects that blur fine structures and high-frequency details, especially at increasing distances from the camera.
    
    \item \textbf{Noise}: Originates from sensor limitations in low-light underwater conditions and from optical effects of suspended particles such as marine snow.
\end{enumerate}

For each degradation type, we design a specialized decoder with architectural inductive biases that naturally guide it toward addressing its target degradation without requiring explicit degradation-specific supervision. Algorithm \ref{alg:specialized_decoder} outlines the general structure of our specialized decoders.

\begin{algorithm}[t]
\caption{Specialized Restoration Decoder}
\label{alg:specialized_decoder}
\begin{algorithmic}
\REQUIRE Hierarchical features $\mathbf{F} = \{\mathbf{f}_0, \mathbf{f}_1, ..., \mathbf{f}_N\}$
\ENSURE Restoration hypothesis $\mathbf{H} \in \mathbb{R}^{3 \times H \times W}$
\STATE $\mathbf{x} \leftarrow \mathbf{f}_N$ \COMMENT{Start with bottleneck features}
\FOR{$i = N-1$ down to $0$}
    \STATE $\mathbf{x} \leftarrow \text{Upsample}(\mathbf{x})$ \COMMENT{2× bilinear upsampling}
    \STATE $\mathbf{x} \leftarrow \text{Concatenate}(\mathbf{x}, \mathbf{f}_i)$ \COMMENT{Skip connection}
    \STATE $\mathbf{x} \leftarrow \phi_i(\mathbf{x})$ \COMMENT{Channel calibration convolution}
    \STATE $\mathbf{x} \leftarrow \text{SpecializedProcessing}_i(\mathbf{x})$ \COMMENT{Decoder-specific processing}
\ENDFOR
\STATE $\mathbf{x} \leftarrow \phi_{\text{out}}(\mathbf{x})$ \COMMENT{Output convolution}
\STATE $\mathbf{H} \leftarrow (\tanh(\mathbf{x}) + 1) / 2$ \COMMENT{Range normalization to [0,1]}
\RETURN $\mathbf{H}$
\end{algorithmic}
\end{algorithm}

The specialized processing functions for each decoder are designed with inductive biases tailored to their target degradation types:

\paragraph{Color Restoration Decoder.} The color decoder implements channel-wise attention mechanisms at multiple scales to recalibrate color-specific features. For feature map $\mathbf{x} \in \mathbb{R}^{C \times h \times w}$ at scale $i$, the specialized processing computes:

\begin{equation}
\text{SpecializedProcessing}_i(\mathbf{x}) = \mathbf{x} \odot \sigma(W_{2,i}(\delta(W_{1,i}(\mathbf{x}))))
\end{equation}

\noindent where $W_{1,i}: \mathbb{R}^C \rightarrow \mathbb{R}^{C/r}$ and $W_{2,i}: \mathbb{R}^{C/r} \rightarrow \mathbb{R}^C$ are learned projections with reduction ratio $r = \max(4, C/16)$. This squeeze-and-excitation attention structure excels at channel recalibration, making it particularly effective for addressing wavelength-dependent color distortions that affect different color channels non-uniformly.

\paragraph{Contrast Enhancement Decoder.} The contrast decoder uses residual blocks designed to enhance local contrast while preserving structural information:

\begin{equation}
\text{SpecializedProcessing}_i(\mathbf{x}) = \mathbf{x} + \phi_{c,i}(\mathbf{x})
\end{equation}

\noindent where $\phi_{c,i}$ consists of convolutional layers with normalization and non-linear activation. This architecture is particularly effective at countering the haze and contrast reduction caused by light scattering, as it can amplify local contrast variations without introducing artifacts.

\paragraph{Detail Recovery Decoder.} The detail decoder employs cascaded residual blocks specifically designed for high-frequency information recovery:

\begin{equation}
\text{SpecializedProcessing}_i(\mathbf{x}) = \mathbf{x} + \sum_{j=1}^{L} \phi_{d,i,j}(\mathbf{x})
\end{equation}

\noindent where $L=2$ and each $\phi_{d,i,j}$ represents a residual block with edge-preserving operations. This cumulative enhancement approach is particularly suited for recovering fine details lost through medium absorption, as it progressively refines high-frequency components while maintaining structural coherence.

\paragraph{Denoising Decoder.} The noise decoder utilizes group convolutions to exploit local correlations while preserving structural information:

\begin{equation}
\text{SpecializedProcessing}_i(\mathbf{x}) = \text{Conv}_{1 \times 1}(\text{GConv}_{3 \times 3}(\phi_{n,i}(\mathbf{x})))
\end{equation}

\noindent where $\text{GConv}_{3 \times 3}$ denotes group convolution with $g = \min(\lfloor C/8 \rfloor, C)$ groups, allowing specialized filtering within channel subgroups, followed by a $1 \times 1$ convolution for cross-group information exchange. This architecture is particularly effective for handling the noise from suspended particles while preserving important image structures.

\subsection{Fusion and Progressive Refinement Mechanisms}
\label{subsec:fusion_refinement}

Our framework incorporates two fusion mechanisms: adaptive hypothesis fusion in the first stage and safety-gated refinement fusion in the second stage. Additionally, we implement a differential degradation analysis for targeted refinement.

\paragraph{Adaptive Hypothesis Fusion.} The base model combines multiple restoration hypotheses $\{H_1, H_2, ..., H_K\}$ using degradation-guided weights. We implement three fusion variants, with learned fusion being our primary approach. The direct fusion establishes a one-to-one mapping between degradation maps and hypotheses:

\begin{equation}
\hat{\mathbf{J}}_1 = \sum_{k=1}^{K} M_k \odot H_k
\end{equation}

\noindent assuming $K$ degradation types correspond directly to $K$ hypotheses. In contrast, our learned fusion approach implements a non-linear mapping from degradation maps to fusion weights:

\begin{equation}
\mathbf{W} = \text{softmax}(f_W(\mathbf{M}))
\end{equation}

\noindent where $f_W$ consists of convolutional layers with residual connections, and the softmax ensures that weights sum to 1 at each spatial location. The fused output is then:

\begin{equation}
\hat{\mathbf{J}}_1 = \sum_{k=1}^{K} W_k \odot H_k
\end{equation}

This learned mapping enables more complex relationships between degradation characteristics and restoration strategies, adapting to various underwater scenes.

\paragraph{Residual Degradation Analysis.} The second stage begins with analyzing residual degradations after initial restoration. We formulate this as a difference-guided estimation process that takes both the original degraded image $\mathbf{I}$ and initial restoration $\hat{\mathbf{J}}_1$ as inputs:

\begin{equation}
\mathbf{M}_r = \mathcal{D}_2(\mathbf{I}, \hat{\mathbf{J}}_1)
\end{equation}

The residual estimator $\mathcal{D}_2$ first computes an absolute difference map $\mathbf{D} = |\mathbf{I} - \hat{\mathbf{J}}_1|$ that highlights regions with significant changes. A difference enhancement module processes this map to create an attention signal:

\begin{equation}
\mathbf{A} = \sigma(f_d(\mathbf{D}))
\end{equation}

\noindent where $f_d$ consists of convolutional layers that extract meaningful patterns from the difference map. The concatenated input $[\mathbf{I}, \hat{\mathbf{J}}_1]$ is processed through an encoder-decoder network whose features are modulated by the attention signal:

\begin{equation}
\mathbf{z} = f_r([\mathbf{I}, \hat{\mathbf{J}}_1]) \odot (1 + \alpha \cdot \mathbf{A})
\end{equation}

\noindent where $f_r$ is the residual estimation network and $\alpha$ is a learnable parameter that controls the influence of the difference enhancement. The final residual degradation maps are obtained by applying a $1 \times 1$ convolution followed by sigmoid activation:

\begin{equation}
\mathbf{M}_r = \sigma(W_r(\mathbf{z}))
\end{equation}

\paragraph{Expert-guided Refinement.} For each residual degradation type, we design a specialized refinement expert that generates targeted corrections. Algorithm \ref{alg:refinement_expert} outlines the refinement process for each expert.

\begin{algorithm}[t]
\caption{Expert-guided Refinement Generation}
\label{alg:refinement_expert}
\begin{algorithmic}
\REQUIRE Original image $\mathbf{I}$, initial restoration $\hat{\mathbf{J}}_1$, expert type $k$
\ENSURE Correction term $\mathbf{C}_k$
\STATE $\mathbf{x} \leftarrow \text{Concatenate}(\mathbf{I}, \hat{\mathbf{J}}_1)$ \COMMENT{6-channel input}
\STATE $\mathbf{x} \leftarrow \phi_{\text{init},k}(\mathbf{x})$ \COMMENT{Initial feature extraction}
\STATE $\mathbf{x} \leftarrow \phi_{\text{res1},k}(\mathbf{x}) + \mathbf{x}$ \COMMENT{First residual block}
\STATE $\mathbf{x} \leftarrow \phi_{\text{res2},k}(\mathbf{x}) + \mathbf{x}$ \COMMENT{Second residual block}
\IF{$k = \text{color}$}
    \STATE $\mathbf{x} \leftarrow \phi_{\text{color}}(\mathbf{x})$ \COMMENT{Color-specific processing}
\ELSIF{$k = \text{contrast}$}
    \STATE $\mathbf{x} \leftarrow \phi_{\text{contrast}}(\mathbf{x})$ \COMMENT{Contrast-specific processing}
\ELSIF{$k = \text{detail}$}
    \STATE $\mathbf{x} \leftarrow \phi_{\text{detail}}(\mathbf{x})$ \COMMENT{Detail-specific processing}
\ELSIF{$k = \text{noise}$}
    \STATE $\mathbf{x} \leftarrow \phi_{\text{noise}}(\mathbf{x})$ \COMMENT{Noise-specific processing}
\ENDIF
\STATE $\mathbf{C}_k \leftarrow \tanh(\mathbf{x}) \cdot \sigma(s_k)$ \COMMENT{Apply adaptive scaling}
\RETURN $\mathbf{C}_k$
\end{algorithmic}
\end{algorithm}

Each expert implements specialized processing tailored to its degradation type:

\begin{itemize}
    \item \textbf{Color refinement expert} focuses on correcting residual color distortions through sequential convolutions that maintain spatial dimensions while refining color relationships.
    
    \item \textbf{Contrast refinement expert} enhances local contrast through a combination of standard convolutions and a residual block that preserves structural integrity.
    
    \item \textbf{Detail refinement expert} employs a more complex pathway with channel expansion and multiple residual connections:
    \begin{equation}
    \phi_{\text{detail}}(\mathbf{x}) = \text{Conv}_{1 \times 1}(\phi_3(\phi_2(\phi_1(\mathbf{x}) + \mathbf{x}) + \phi_1(\mathbf{x})) + \phi_2(\phi_1(\mathbf{x})))
    \end{equation}
    where $\phi_1$, $\phi_2$, and $\phi_3$ are convolutional layers with normalization and activation.
    
    \item \textbf{Noise refinement expert} utilizes group convolutions with multiple paths:
    \begin{equation}
    \phi_{\text{noise}}(\mathbf{x}) = \text{Conv}_{1 \times 1}(\text{GConv}_{3 \times 3}(\phi_n(\mathbf{x})))
    \end{equation}
    with groups $g = \min(C/4, C)$ to process channel subsets independently.
\end{itemize}

Each expert produces a correction term $\mathbf{C}_k$ with adaptive scaling controlled by a learnable parameter $s_k$. The scaling ensures that corrections remain moderate and prevents over-correction.

\paragraph{Safety-gated Fusion.} The individual correction terms are combined through a residual degradation-guided fusion with an additional safety mechanism:

\begin{equation}
\mathbf{W}_r = \text{softmax}(f_r(\mathbf{M}_r))
\end{equation}

\noindent where $f_r$ is a mapping function consisting of convolutional layers that transform residual degradation maps into fusion weights. The fused correction is computed as:

\begin{equation}
\mathbf{C}' = \sum_{k=1}^{K} (W_r)_k \odot \mathbf{C}_k
\end{equation}

To prevent unnecessary corrections in well-restored regions, we implement a safety gate that examines the initial restoration:

\begin{equation}
G(\hat{\mathbf{J}}_1) = \sigma(f_g(\hat{\mathbf{J}}_1))
\end{equation}

\noindent where $f_g$ consists of convolutional layers that identify regions requiring minimal correction. The safety gate produces a spatial mask $G \in \mathbb{R}^{1 \times H \times W}$ with values in $[0,1]$, where lower values indicate well-restored regions. The final correction is modulated by this gate and a global scaling factor:

\begin{equation}
\mathbf{C} = \mathbf{C}' \odot G(\hat{\mathbf{J}}_1) \cdot \sigma(s_g)
\end{equation}

\noindent where $s_g$ is a learnable parameter controlling the overall refinement intensity. The final restoration is obtained by:

\begin{equation}
\hat{\mathbf{J}} = \text{clamp}(\hat{\mathbf{J}}_1 + \mathbf{C}, 0, 1)
\end{equation}

\noindent ensuring that the pixel values remain within the valid range.

\subsection{Specialization Through Architecture and Loss Functions}
\label{subsec:specialization}

A key aspect of our approach is how specialization emerges without explicit supervision for each degradation type. This is achieved through a combination of architectural inductive biases and specialized loss functions.

\paragraph{Architectural Inductive Biases.} Each decoder and refinement expert incorporates architectural elements specifically designed for its target degradation type:

\begin{itemize}
    \item The color restoration decoder's channel attention mechanism is particularly effective for addressing wavelength-dependent color distortions because it can selectively recalibrate different color channels with varying intensities.
    
    \item The contrast enhancement decoder's residual blocks are designed to amplify local contrast variations while preserving structural integrity, making them well-suited for addressing the haze and low contrast caused by light scattering.
    
    \item The detail recovery decoder's cascaded residual structure with multiple pathways excels at recovering high-frequency information lost through medium absorption and forward scattering.
    
    \item The denoising decoder's group convolutions enable specialized filtering within channel subgroups followed by cross-group information exchange, which is particularly effective for removing noise while preserving important image structures.
\end{itemize}

These architectural choices provide implicit guidance for each decoder to specialize in its target degradation type, even without explicit supervision signals.

\paragraph{Specialized Loss Functions.} We reinforce this specialization through carefully designed loss functions during training:

\begin{itemize}
    \item \textbf{Diversity Loss:} By explicitly penalizing similarity between different hypotheses, this loss encourages each decoder to develop unique restoration capabilities:
    \begin{equation}
    \mathcal{L}_{div} = \frac{1}{K(K-1)/2} \sum_{i=1}^{K} \sum_{j=i+1}^{K} \text{cos\_sim}(H_i, H_j)
    \end{equation}
    where $\text{cos\_sim}(H_i, H_j)$ computes the cosine similarity between flattened hypothesis tensors. This loss is minimized when hypotheses are orthogonal, encouraging specialization.
    
    \item \textbf{Degradation Consistency Loss:} This loss ensures that the estimated degradation maps correlate with actual degradations:
    \begin{equation}
    \mathcal{L}_{dc} = \|\text{norm}(\mathbf{M}_{sum}) - \text{norm}(\|\mathbf{I} - \mathbf{J}\|_1)\|_2^2
    \end{equation}
    where $\mathbf{M}_{sum} = \sum_{k=1}^{K} M_k$ is the summed degradation map, and $\text{norm}(\cdot)$ normalizes the values to $[0,1]$ range. This guides the degradation estimator to produce meaningful maps that reflect the spatial distribution and severity of degradations.
\end{itemize}

Through the combination of these architectural inductive biases and specialized loss functions, each decoder naturally develops specialization in addressing its target degradation type without requiring explicit degradation-specific supervision.

\subsection{Loss Functions and Training Strategy}
\label{subsec:training_details}

We employ a multi-stage training strategy with carefully designed loss functions to ensure effective learning at each stage.

\paragraph{Base Model Training.} The first stage trains only the base model with a combination of five loss components:

\begin{equation}
\mathcal{L}_1 = \lambda_{\ell_1} \mathcal{L}_{\ell_1} + \lambda_{ssim} \mathcal{L}_{ssim} + \lambda_p \mathcal{L}_p + \lambda_{div} \mathcal{L}_{div} + \lambda_{dc} \mathcal{L}_{dc}
\end{equation}

The reconstruction losses include L1 loss, SSIM loss, and perceptual loss:

\begin{equation}
\mathcal{L}_{\ell_1} = \|\hat{\mathbf{J}}_1 - \mathbf{J}\|_1
\end{equation}

\begin{equation}
\mathcal{L}_{ssim} = 1 - \text{SSIM}(\hat{\mathbf{J}}_1, \mathbf{J})
\end{equation}

\begin{equation}
\mathcal{L}_p = \sum_{l} \lambda_l \|\Phi_l(\hat{\mathbf{J}}_1) - \Phi_l(\mathbf{J})\|_1
\end{equation}

\noindent where $\mathbf{J}$ is the ground truth reference image, SSIM is the structural similarity index, and $\Phi_l$ represents the $l$-th layer features of a pre-trained VGG-19 network.

We also apply auxiliary losses to individual hypotheses:

\begin{equation}
\mathcal{L}_{aux} = \frac{1}{K} \sum_{k=1}^{K} \|H_k - \mathbf{J}\|_1
\end{equation}

\noindent encouraging each hypothesis to approximate the reference image, though with less weight than the primary losses.

\paragraph{Refinement Stage Training.} In the second stage, we freeze the base model parameters and train only the refinement components using:

\begin{equation}
\mathcal{L}_2 = \mathcal{L}_{recon} + \lambda_{mag} \mathcal{L}_{mag} + \lambda_{imp} \mathcal{L}_{imp}
\end{equation}

The reconstruction loss $\mathcal{L}_{recon}$ combines L1, SSIM, and perceptual losses for the final output $\hat{\mathbf{J}}$, similar to the first stage. The magnitude loss penalizes excessive correction:

\begin{equation}
\mathcal{L}_{mag} = \|\mathbf{C}\|_1
\end{equation}

\noindent encouraging minimal refinement where possible. The progressive improvement loss explicitly requires the final output to be better than the initial restoration:

\begin{equation}
\mathcal{L}_{imp} = \max(0, \|\hat{\mathbf{J}} - \mathbf{J}\|_1 - \|\hat{\mathbf{J}}_1 - \mathbf{J}\|_1 + \epsilon)
\end{equation}

\noindent where $\epsilon = 0.01$ is a small constant that encourages a minimum level of improvement. This loss is activated only when the final restoration has higher error than the initial restoration, penalizing cases where refinement degrades quality.

\paragraph{Training Strategy.} We implement a two-phase training strategy:

1. \textbf{Base Model Training}: We train the degradation estimation, hypothesis generation, and fusion components for 300 epochs using the Adam optimizer with an initial learning rate of $10^{-4}$ and cosine annealing schedule.

2. \textbf{Refinement Training}: With the base model frozen, we train the residual estimation and refinement components for 100 epochs with a reduced learning rate of $5 \times 10^{-5}$.

This staged approach ensures stable learning, with each component trained to fulfill its specific role in the restoration pipeline. During inference, the model processes images in a single forward pass through both stages, with negligible additional computational cost for the refinement stage compared to single-stage approaches.

\section{Experimental Setup and Implementation Details}
\label{sec:exp_setup}

\subsection{Training Configuration}

We implement our TIDE framework in PyTorch and conduct all experiments on a single NVIDIA RTX 4070 Ti SUPER GPU. For network optimization, we employ a multi-stage training strategy with carefully tuned hyperparameters. In the base model training phase, we use the Adam optimizer with an initial learning rate of $1 \times 10^{-4}$, weight decay of $1 \times 10^{-4}$, and train for 300 epochs with a batch size of 16. We apply cosine annealing with warm restarts for the learning rate schedule, with cycle length of 50 epochs. During the refinement stage training, we freeze the base model parameters and train the refinement components for 100 epochs using a reduced learning rate of $5 \times 10^{-5}$ while maintaining the same optimizer configuration. 

We determine the optimal loss function weights through a grid search for hyperparameter tuning. During base model training, we use $\lambda_{\ell_1}=1.0$, $\lambda_{ssim}=0.1$, $\lambda_{perceptual}=0.1$, $\lambda_{diversity}=0.05$, and $\lambda_{degradation}=0.1$. In the refinement stage, we set $\lambda_{recon}=1.0$, $\lambda_{magnitude}=0.1$, and $\lambda_{improve}=0.5$. For the progressive combined loss during fine-tuning, we balance the base and refinement objectives with $\lambda_{base}=0.7$ and $\lambda_{refinement}=1.0$. We apply gradient clipping with a threshold of 1.0 during all training phases to ensure stable convergence. Mixed precision training is employed to improve computational efficiency while maintaining numerical stability.

\subsection{Network Architecture}

The feature extraction backbone consists of 5 downsampling layers starting with 64 base channels and doubling the channel count at each downsampling step up to a maximum of 512 channels. The degradation estimation module employs a lightweight structure with 32 base channels and 4 degradation types corresponding to color distortion, contrast reduction, detail loss, and noise. For normalization, we use instance normalization throughout the network, and LeakyReLU with slope 0.2 as the activation function. The specialized decoders share the same general architecture but incorporate domain-specific processing blocks as detailed in Section~\ref{sec:extended_method}.

The fusion mechanism implements our learned mapping approach by default, though we also explore a direct fusion alternative in our ablation studies. The residual degradation estimator uses a condensed encoder-decoder structure with 2 downsampling layers, while each refinement expert employs adaptive scaling parameters initialized to 0.1 to ensure gradual refinement during early training. The safety gate in the refinement fusion module uses 16 intermediate channels to generate the spatial modulation mask.

\subsection{Dataset Details}
\label{subsec:dataset_details}

We conduct experiments on three widely-used underwater image datasets for training and standard evaluation: UIEB~\cite{li2019underwater}, EUVP~\cite{islam2020fast}, and SUIM-E~\cite{qi2022sguie}. Additionally, we include the RUIE dataset~\cite{turbidDataset} specifically to evaluate performance on turbid water conditions.

\paragraph{Standard Benchmark Datasets.} The EUVP dataset provides 11,435 paired images for training and 515 pairs for testing, all at the same resolution. From the UIEB dataset, which contains 890 paired images, we follow the protocol in~\cite{li2019underwater} by selecting 800 images for training and using the remaining 90 for testing. The SUIM-E dataset consists of 1,635 images, where 1,525 images are used for training and 110 are reserved for evaluation, following the procedure in~\cite{ms2_litnet}.

\paragraph{Turbid Water Evaluation.} To evaluate TIDE under challenging real-world conditions beyond clear-water scenarios, we utilize the RUIE dataset~\cite{turbidDataset}, focusing on its UIQS (Underwater Image Quality Set) and UCCS (Underwater Color Cast Set) subsets. The UIQS subset comprises 3,630 images of real-world underwater scenes systematically grouped into five quality levels (A--E) based on the UCIQE metric, which quantifies image quality using chromaticity, saturation, and contrast measurements. This graded organization enables comprehensive assessment across varying degrees of turbidity and visual degradation. The UCCS subset contains 300 images selected from UIQS, specifically targeting color restoration capabilities. These images are equally divided into three groups - bluish, greenish/yellowish, and blue-green - each corresponding to common underwater color cast conditions encountered in practice.

For turbid water evaluation, we use the model trained on UIEB without any fine-tuning or domain adaptation, allowing us to assess both relative performance and generalization capabilities. This approach tests whether the model can generalize to significantly more challenging underwater environments with optical properties different from those represented in the training data. We compare TIDE against FUSION~\cite{fusion} and LitNet~\cite{ms2_litnet}, the two strongest baselines from our main experiments, using both quantitative metrics (UICM and UIConM) and qualitative visual assessments focused on color correction, contrast enhancement, and detail preservation in turbid conditions.

\subsection{Turbid Water Evaluation Metrics}
\label{subsec:turbid_metrics}

For turbid water evaluation, we utilize metrics specifically designed to assess underwater image quality. These metrics target the unique challenges of turbid water conditions, focusing on color cast correction and contrast enhancement:

\paragraph{Underwater Image Colorfulness Measure (UICM).} This metric quantifies the color quality in underwater images by analyzing the chromatic components in the LAB color space. For an input image $I$, we first convert it to the LAB color space and extract the $a$ and $b$ channels, which represent the green-red and blue-yellow opponent color pairs, respectively. The UICM is then calculated as:

\begin{equation}
\text{UICM}(I) = -0.0268 \cdot \sqrt{\mu_a^2 + \mu_b^2} + 0.1586 \cdot (\sigma_a + \sigma_b)
\end{equation}

\noindent where $\mu_a$ and $\mu_b$ are the means of the $a$ and $b$ channels, while $\sigma_a$ and $\sigma_b$ are their respective standard deviations. This formulation has two key components: the negative term penalizes strong color casts (high mean values in either direction), while the positive term rewards color variation and richness (high standard deviations). Higher UICM values indicate better color balance and diversity, which is particularly important for turbid underwater scenes where color distortion is severe.

\begin{figure}[t]
    \centering
    
    \begin{subfigure}[t]{0.23\columnwidth}
        \centering
        \includegraphics[width=\linewidth]{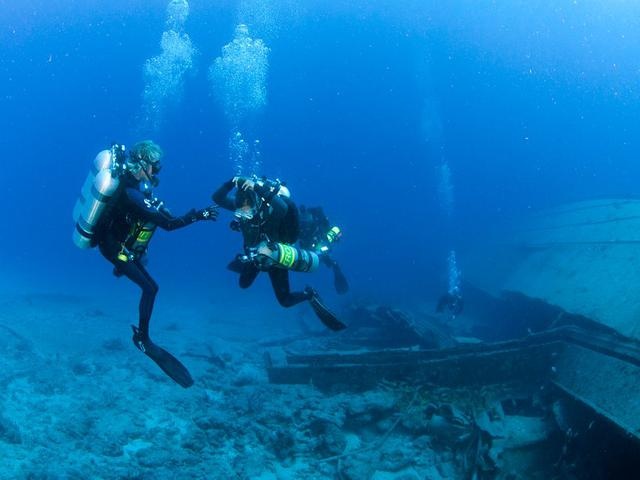}
        \caption*{\small Original}
    \end{subfigure}
    \hfill
    \begin{subfigure}[t]{0.23\columnwidth}
        \centering
        \begin{tikzpicture}
            \node[anchor=south west,inner sep=0] (image) at (0,0) {\includegraphics[width=\linewidth]{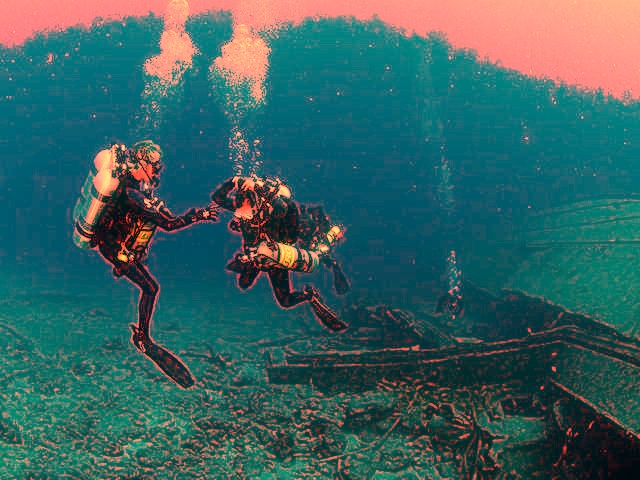}};
            \node[anchor=north east, fill=white, fill opacity=0.5, text opacity=1, inner sep=2pt, font=\tiny\bfseries] at (image.north east) {12.64 dB};
        \end{tikzpicture}
        \caption*{\small WB}
    \end{subfigure}
    \hfill
    \begin{subfigure}[t]{0.23\columnwidth}
        \centering
        \begin{tikzpicture}
            \node[anchor=south west,inner sep=0] (image) at (0,0) {\includegraphics[width=\linewidth]{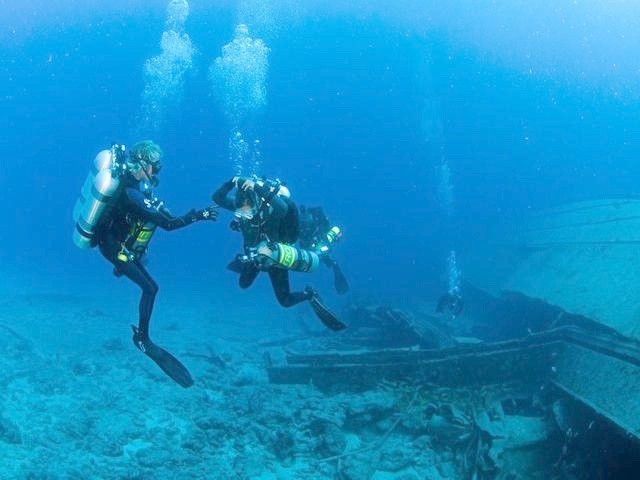}};
            \node[anchor=north east, fill=white, fill opacity=0.5, text opacity=1, inner sep=2pt, font=\tiny\bfseries] at (image.north east) {10.05 dB};
        \end{tikzpicture}
        \caption*{\small GC}
    \end{subfigure}
    \hfill
    \begin{subfigure}[t]{0.23\columnwidth}
        \centering
        \begin{tikzpicture}
            \node[anchor=south west,inner sep=0] (image) at (0,0) {\includegraphics[width=\linewidth]{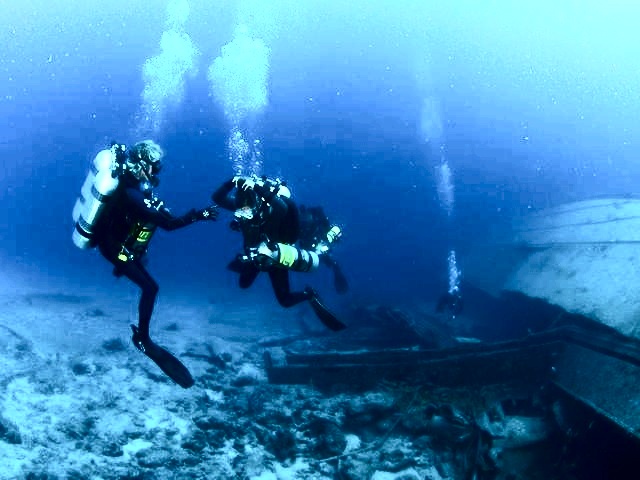}};
            \node[anchor=north east, fill=white, fill opacity=0.5, text opacity=1, inner sep=2pt, font=\tiny\bfseries] at (image.north east) {13.26 dB};
        \end{tikzpicture}
        \caption*{\small HE}
    \end{subfigure}
    
    \vspace{1mm}
    \begin{subfigure}[t]{0.23\columnwidth}
        \centering
        \begin{tikzpicture}
            \node[anchor=south west,inner sep=0] (image) at (0,0) {\includegraphics[width=\linewidth]{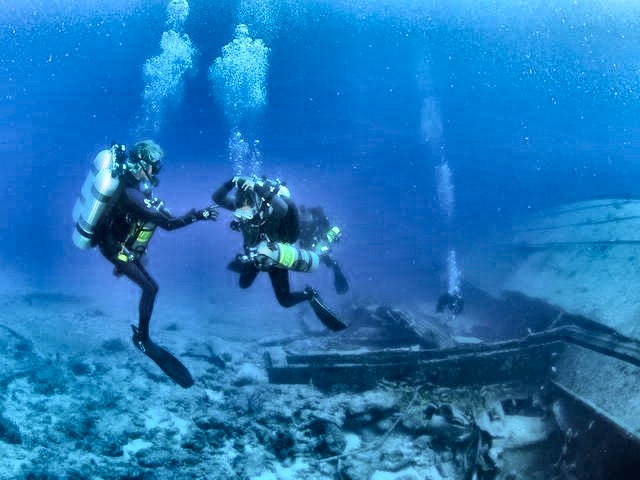}};
            \node[anchor=north east, fill=white, fill opacity=0.5, text opacity=1, inner sep=2pt, font=\tiny\bfseries] at (image.north east) {12.43 dB};
        \end{tikzpicture}
        \caption*{\small CLAHE}
    \end{subfigure}
    \hfill
    \begin{subfigure}[t]{0.23\columnwidth}
        \centering
        \begin{tikzpicture}
            \node[anchor=south west,inner sep=0] (image) at (0,0) {\includegraphics[width=\linewidth]{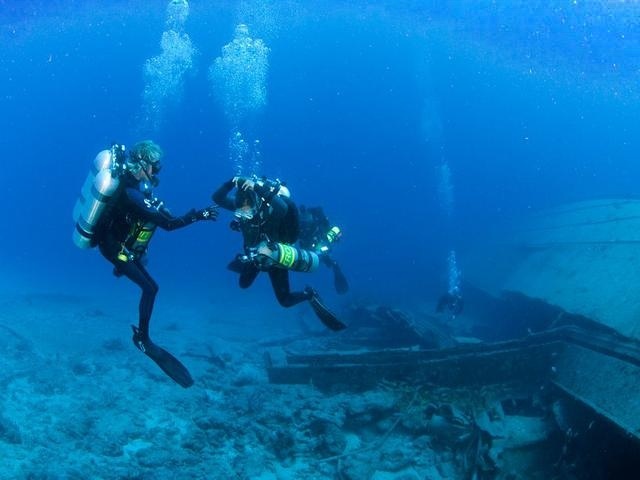}};
            \node[anchor=north east, fill=white, fill opacity=0.5, text opacity=1, inner sep=2pt, font=\tiny\bfseries] at (image.north east) {10.78 dB};
        \end{tikzpicture}
        \caption*{\small DCP}
    \end{subfigure}
    \hfill
    \begin{subfigure}[t]{0.23\columnwidth}
        \centering
        \begin{tikzpicture}
            \node[anchor=south west,inner sep=0] (image) at (0,0) {\includegraphics[width=\linewidth]{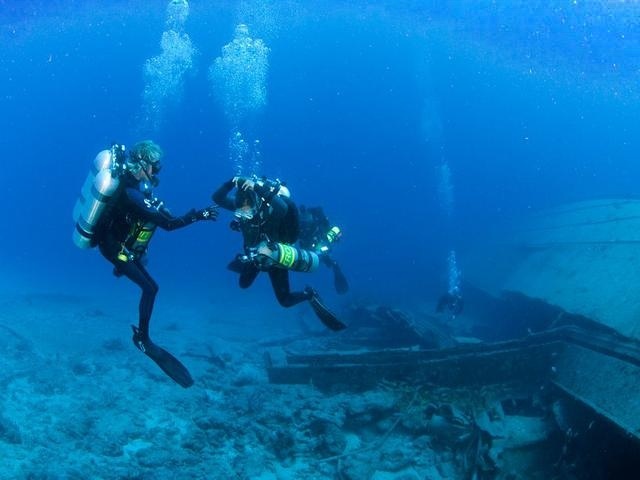}};
            \node[anchor=north east, fill=white, fill opacity=0.5, text opacity=1, inner sep=2pt, font=\tiny\bfseries] at (image.north east) {10.78 dB};
        \end{tikzpicture}
        \caption*{\small UDCP}
    \end{subfigure}
    \hfill
    \begin{subfigure}[t]{0.23\columnwidth}
        \centering
        \begin{tikzpicture}
            \node[anchor=south west,inner sep=0] (image) at (0,0) {\includegraphics[width=\linewidth]{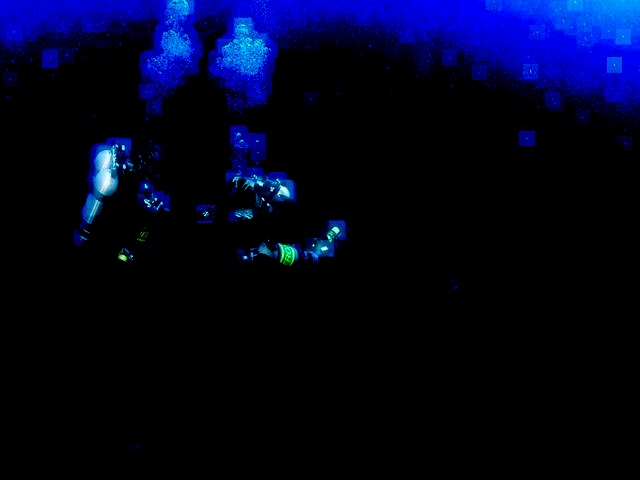}};
            \node[anchor=north east, fill=white, fill opacity=0.5, text opacity=1, inner sep=2pt, font=\tiny\bfseries] at (image.north east) {6.98 dB};
        \end{tikzpicture}
        \caption*{\small RCP}
    \end{subfigure}
    
    \caption{Comparison between traditional UIR methods. Abbreviations: WB - White Balance, GC - Gamma Correction (1.5), HE - Histogram Equalization, CLAHE - Contrast Limited Adaptive Histogram Equalization, DCP - Dark Channel Prior, UDCP - Underwater Dark Channel Prior, RCP - Red Channel Prior.}
    \label{fig:method_comparison}
\end{figure}

\paragraph{Underwater Image Contrast Measure (UIConM).} This metric evaluates local contrast enhancement, which is crucial for visibility in turbid waters. For an image converted to grayscale $G$, UIConM analyzes the local standard deviations across the entire image:

\begin{equation}
\text{UIConM}(I) = \frac{1}{HW} \sum_{i=1}^{H} \sum_{j=1}^{W} \sigma_{ij}
\end{equation}

\noindent where $\sigma_{ij}$ is the local standard deviation in a $5 \times 5$ window centered at pixel position $(i,j)$, calculated as:

\begin{equation}
\sigma_{ij} = \sqrt{\frac{1}{N} \sum_{(p,q) \in W_{ij}} (G(p,q) - \mu_{ij})^2}
\end{equation}

\noindent where $W_{ij}$ is the $5 \times 5$ window around pixel $(i,j)$, $N=25$ is the window size, and $\mu_{ij}$ is the local mean in the window. Higher UIConM values indicate better local contrast enhancement, which improves visibility of structures and details in turbid underwater environments.

Together, these metrics provide a comprehensive assessment of color correction and contrast enhancement capabilities, which are particularly challenging in turbid water conditions. They complement the standard metrics (PSNR, SSIM, LPIPS) used in our main experiments by specifically targeting the unique challenges of underwater image restoration in turbid environments.

\begin{figure*}[t]
    \centering
    \begin{subfigure}{\textwidth}
        \centering
        \includegraphics[width=\textwidth]{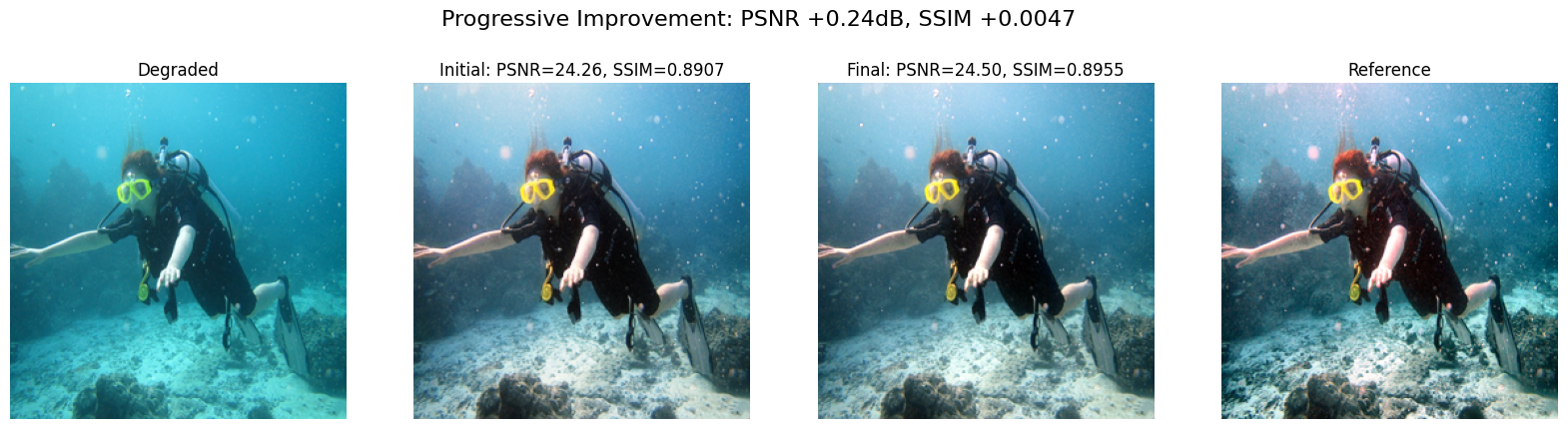}
        \caption{Progressive improvement visualization showing the diver scene. From left to right: degraded input, initial restoration (PSNR=24.26, SSIM=0.8907), final restoration (PSNR=24.50, SSIM=0.8955), and reference image. Our refinement stage achieves a +0.24dB PSNR and +0.0047 SSIM improvement over the initial restoration.}
        \label{fig:example1_progression}
    \end{subfigure}
    
    \vspace{0.5em}
    
    \begin{subfigure}{\textwidth}
        \centering
        \includegraphics[width=\textwidth]{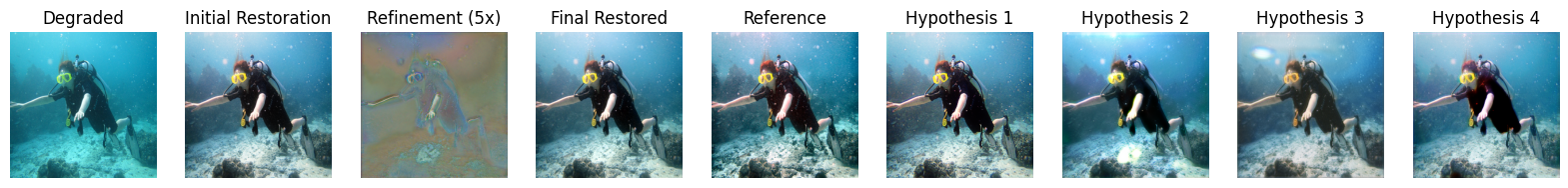}
        \caption{Complete restoration pipeline for the diver scene. From left to right: degraded input, initial restoration, refinement visualization (5× amplified for visibility), final restored result, reference image, and the four specialized hypotheses (color, contrast, detail, and denoising).}
        \label{fig:example1_hypotheses}
    \end{subfigure}
    
    \vspace{0.5em}
    
    \begin{subfigure}{\textwidth}
        \centering
        \includegraphics[width=\textwidth]{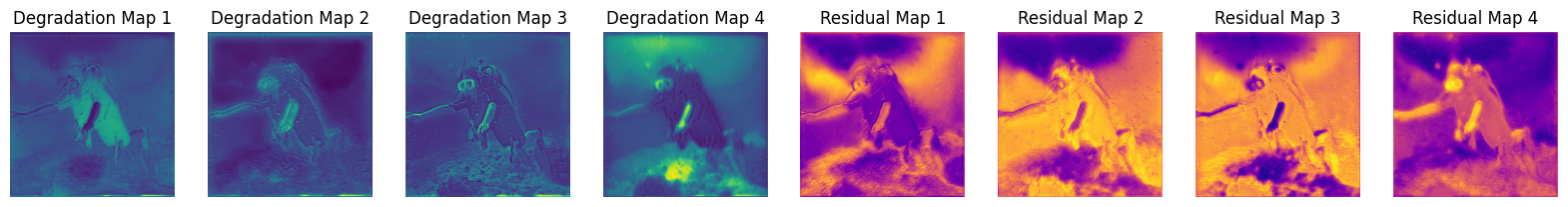}
        \caption{Degradation and residual maps for the diver scene. Left four images: the estimated degradation maps from the first stage, highlighting color distortion, contrast reduction, detail loss, and noise. Right four images: the residual degradation maps from the second stage, indicating regions requiring additional refinement.}
        \label{fig:example1_maps}
    \end{subfigure}
    
    \caption{Comprehensive analysis of the diver scene. The degraded image suffers from greenish color cast and reduced contrast. The initial restoration significantly improves color balance, while the refinement stage enhances detail visibility in the diver's equipment and seafloor textures. The degradation maps clearly identify the upper region as having more severe color distortion, while the residual maps focus refinement efforts on the diver and foreground elements.}
    \label{fig:example1_full}
\end{figure*}

\section{Additional Experiments}
\label{sec:additional_exp}

\subsection{Comparison with Traditional Methods}

We compare our TIDE framework against several classical image processing techniques commonly used for underwater image enhancement. As shown in Figure~\ref{fig:method_comparison}, traditional methods exhibit notable limitations. White Balance adjusts RGB channels using the gray-world assumption but fails to account for wavelength-dependent attenuation, leading to unnatural colors. Gamma Correction (1.5) boosts brightness but doesn’t correct color distortion or adapt to local degradation. Histogram Equalization improves global contrast yet often amplifies noise and causes oversaturation, while CLAHE offers better local contrast but still enhances noise and fails to address color casts. Dark Channel Prior, designed for land-based dehazing, performs poorly underwater due to its invalid assumptions in bluish environments. Its underwater variant accounts for wavelength attenuation but still leaves a bluish tint. The Red Channel Prior enhances the lost red component but tends to overcompensate, resulting in unrealistic reddish tones.

\subsection{Extended Hardware Efficiency Analysis}
\label{subsec:hardware_efficiency}

We provide a detailed analysis of TIDE's computational efficiency to demonstrate its viability for both real-time applications and high-throughput batch processing. Our architecture achieves this efficiency through thoughtful design choices that minimize overhead despite the sophisticated two-stage approach.

\begin{table}[t]
\centering
\caption{Hardware efficiency analysis on RTX 4070 Ti SUPER showing the minimal overhead of the refinement stage.}
\label{tab:hardware_efficiency}
\resizebox{\columnwidth}{!}{%
\begin{tabular}{lcccc}
\toprule
\multicolumn{5}{c}{\textbf{Model Architecture}} \\
\midrule
Component & Parameters (M) & Size (MB) & Overhead & Memory (GB) \\
Base Model & 114.5 & 436.9 & - & 0.44 \\
Refinement Stage & 1.25 & 4.8 & 1.1\% & +0.0 \\
\textbf{Total Progressive} & \textbf{115.8} & \textbf{441.6} & \textbf{1.1\%} & \textbf{0.44} \\
\bottomrule
\end{tabular}%
}
\end{table}

\begin{figure}[t]
    \centering
    \includegraphics[width=\columnwidth]{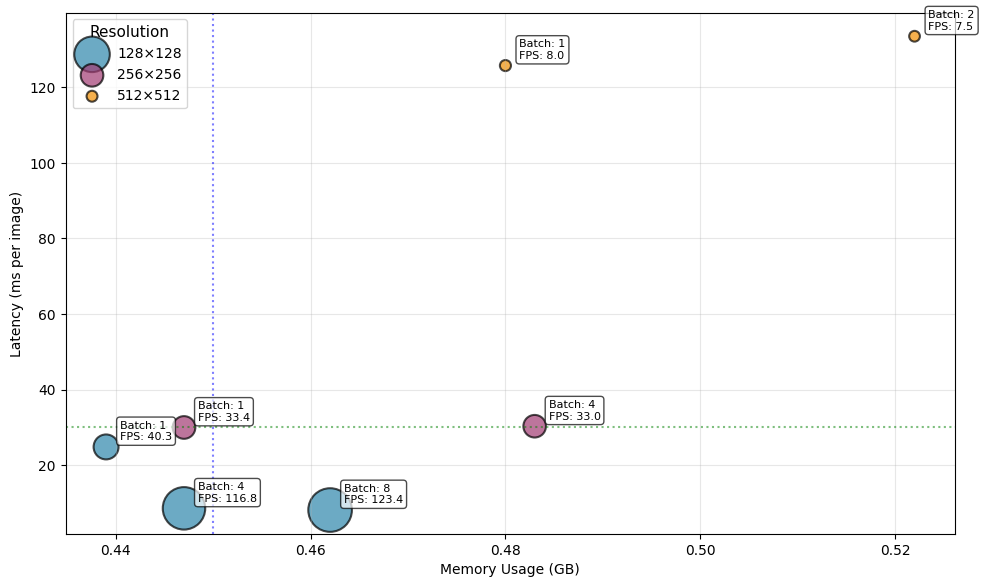}
    \caption{Memory vs. latency analysis across different resolutions and batch sizes. Bubble size corresponds to throughput (FPS). The plot demonstrates TIDE's scalability and efficiency, with configurations in the lower left quadrant achieving real-time performance ($>$30 FPS) with minimal memory usage.}
    \label{fig:efficiency}
\end{figure}

Table~\ref{tab:hardware_efficiency} quantifies the minimal computational burden introduced by our refinement stage. Adding only 1.25M parameters (1.1\% overhead), the refinement components generate targeted correction terms rather than full restorations, enabling efficient resource utilization. The runtime memory footprint remains unchanged at 0.44GB when adding the refinement stage, demonstrating effective activation reuse between stages. Figure~\ref{fig:efficiency} visualizes TIDE's performance characteristics across different configurations, highlighting three key efficiency mechanisms:

\paragraph{Parallel Feature Processing.} TIDE's specialized decoders operate on shared features simultaneously rather than sequentially, enabling efficient hypothesis generation. This parallel architecture yields impressive throughput: at 128×128 resolution with batch size 8, the model achieves 123.4 FPS (8.1ms latency) while maintaining a modest 0.46GB memory footprint. Even at 256×256 resolution, our model delivers real-time performance of 33.4 FPS (30.0ms latency) with batch size 1.

\paragraph{Selective Computation via Degradation-Guided Fusion.} By applying computation proportionally to degradation severity, TIDE avoids redundant processing in well-preserved regions. This targeted approach enables excellent batch scaling properties: at 128×128 resolution, throughput increases 3× from 40.3 FPS (batch 1) to 123.4 FPS (batch 8) with only a 0.02GB memory increase. Similarly, at 256×256 resolution, the model maintains consistent real-time performance when scaling from batch size 1 (33.4 FPS) to batch size 4 (33.0 FPS) with minimal memory overhead (0.04GB).

\paragraph{Lightweight Refinement Design.} Our refinement stage achieves efficiency through sparse correction generation rather than full image reprocessing. This approach results in no measurable increase in runtime memory usage while significantly improving restoration quality. The complete performance profile reveals five real-time capable configurations:

\begin{itemize}
    \item 128×128 (batch 1): 40.3 FPS, 24.8ms latency, 0.439GB memory
    \item 128×128 (batch 4): 116.8 FPS, 8.6ms latency, 0.44GB memory
    \item 128×128 (batch 8): 123.4 FPS, 8.1ms latency, 0.46GB memory
    \item 256×256 (batch 1): 33.4 FPS, 30.0ms latency, 0.44GB memory
    \item 256×256 (batch 4): 33.0 FPS, 30.3ms latency, 0.48GB memory
\end{itemize}

For higher resolution applications (512×512), the model maintains consistent throughput of 7.5-8.0 FPS across different batch sizes, suitable for offline processing of high-resolution underwater imagery. These results demonstrate that TIDE's sophisticated degradation modeling and two-stage approach achieve superior restoration quality without imposing prohibitive computational requirements, making it viable for deployment across a spectrum of hardware configurations and application scenarios.

\subsection{Extended Qualitative Analysis}

To further demonstrate the effectiveness of our two-stage approach, we present additional qualitative results across different datasets. These examples showcase TIDE's ability to handle diverse underwater degradations and provide a comprehensive view of our restoration pipeline's internal workings.

\begin{figure*}[t]
    \centering
    \begin{subfigure}{\textwidth}
        \centering
        \includegraphics[width=\textwidth]{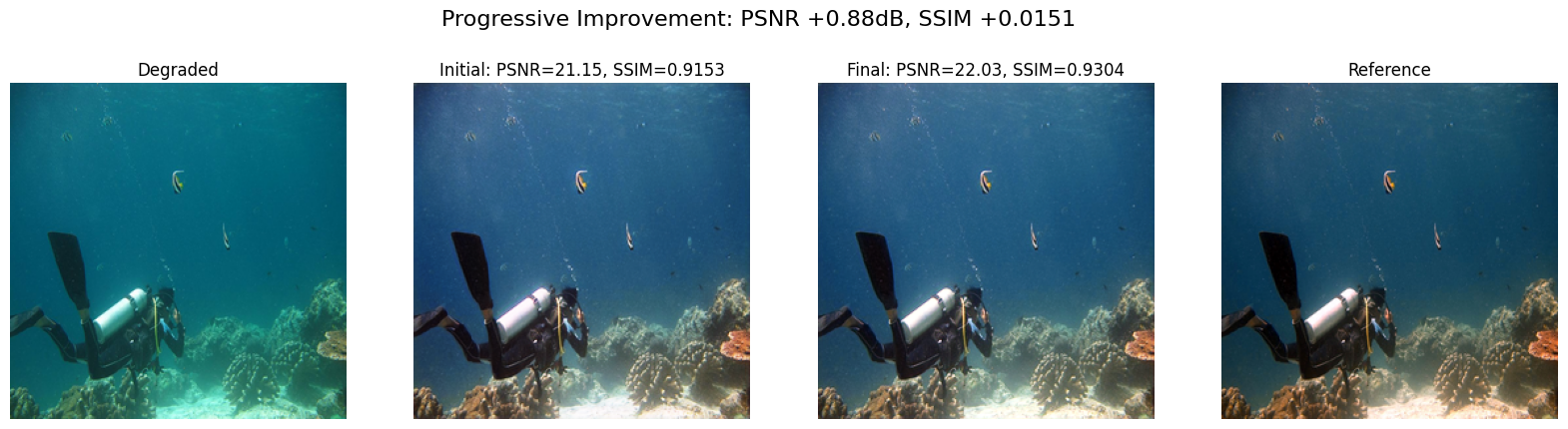}
        \caption{Progressive improvement visualization showing an underwater scene with coral formations. From left to right: degraded input, initial restoration (PSNR=21.15, SSIM=0.9153), final restoration (PSNR=22.03, SSIM=0.9304), and reference image. Our refinement stage achieves a significant +0.88dB PSNR and +0.0151 SSIM improvement.}
        \label{fig:example2_progression}
    \end{subfigure}
    
    \vspace{0.5em}
    
    \begin{subfigure}{\textwidth}
        \centering
        \includegraphics[width=\textwidth]{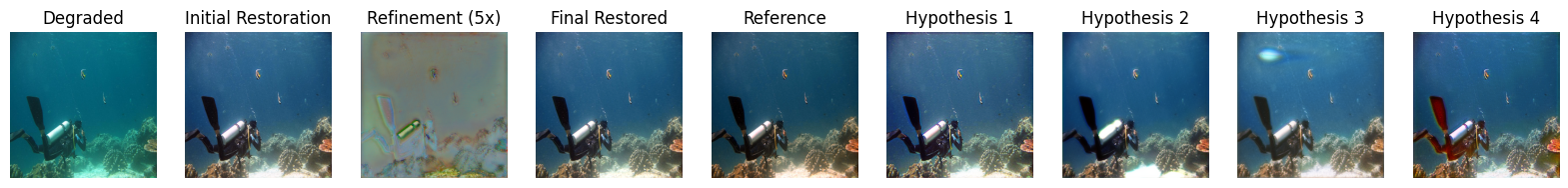}
        \caption{Complete restoration pipeline for the coral scene. From left to right: degraded input, initial restoration, refinement visualization (5× amplified), final restored result, reference image, and the four specialized hypotheses (color, contrast, detail, and denoising).}
        \label{fig:example2_hypotheses}
    \end{subfigure}
    
    \vspace{0.5em}
    
    \begin{subfigure}{\textwidth}
        \centering
        \includegraphics[width=\textwidth]{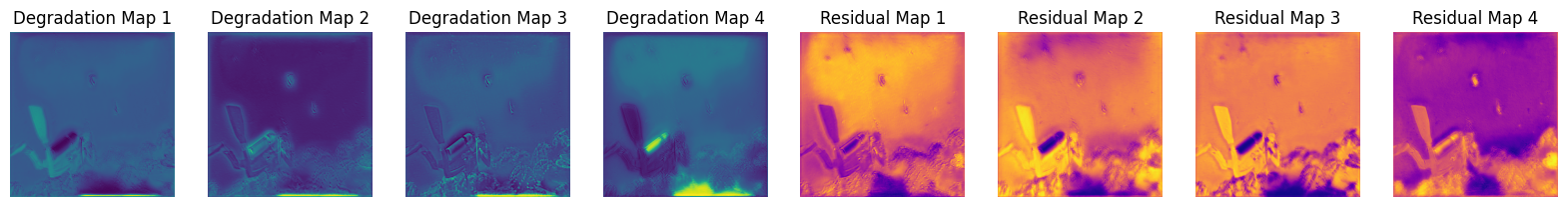}
        \caption{Degradation and residual maps for the coral scene. The degradation maps (left four) show strong color distortion in the water column and significant detail loss in the coral region. The residual maps (right four) indicate that refinement focuses on enhancing the diver's equipment and the coral textures.}
        \label{fig:example2_maps}
    \end{subfigure}
    
    \caption{Comprehensive analysis of the coral scene. This example demonstrates TIDE's ability to handle scenes with complex foreground structures and varying degradation characteristics. The degradation maps indicate stronger color distortion in the water column and noise in the coral regions, while the hypotheses show how each decoder contributes differently to the final restoration.}
    \label{fig:example2_full}
\end{figure*}

\begin{figure*}[t]
    \centering
    \begin{subfigure}{\textwidth}
        \centering
        \includegraphics[width=\textwidth]{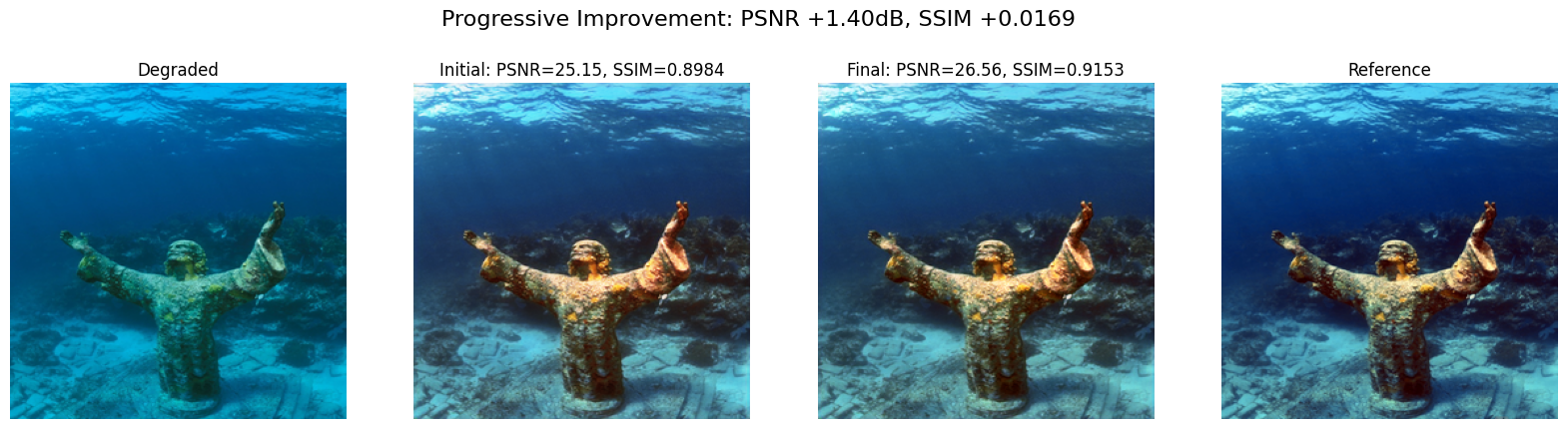}
        \caption{Progressive improvement visualization showing the underwater statue. From left to right: degraded input, initial restoration (PSNR=25.15, SSIM=0.8984), final restoration (PSNR=26.56, SSIM=0.9153), and reference image. The refinement stage achieves a substantial +1.40dB PSNR and +0.0169 SSIM improvement.}
        \label{fig:example3_progression}
    \end{subfigure}
    
    \vspace{0.5em}
    
    \begin{subfigure}{\textwidth}
        \centering
        \includegraphics[width=\textwidth]{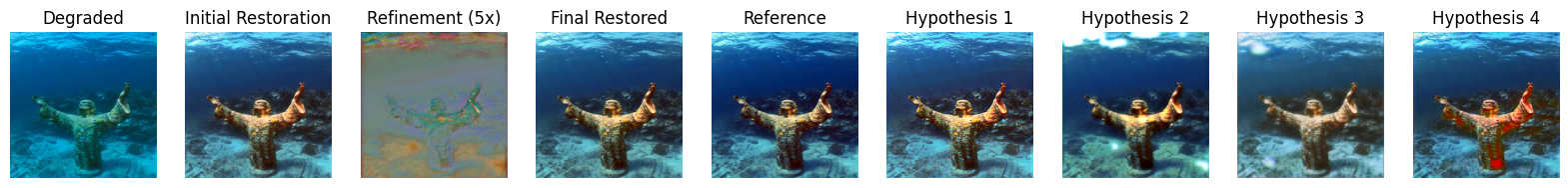}
        \caption{Complete restoration pipeline for the statue scene. From left to right: degraded input, initial restoration, refinement visualization (5× amplified), final restored result, reference image, and the four specialized hypotheses. Note how the color and contrast hypotheses (first two) contribute most to the overall restoration.}
        \label{fig:example3_hypotheses}
    \end{subfigure}
    
    \vspace{0.5em}
    
    \begin{subfigure}{\textwidth}
        \centering
        \includegraphics[width=\textwidth]{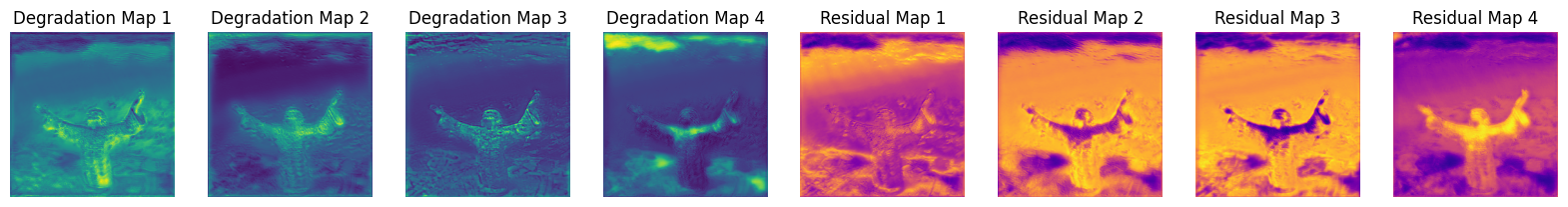}
        \caption{Degradation and residual maps for the statue scene. The degradation maps show stronger color distortion and detail loss around the statue, while the residual maps highlight regions requiring additional refinement, particularly in the statue's textural details and the water surface.}
        \label{fig:example3_maps}
    \end{subfigure}
    
    \caption{Comprehensive analysis of the underwater statue scene. This example demonstrates the largest quantitative improvement among our examples (+1.40dB PSNR), highlighting the effectiveness of our two-stage approach on challenging scenes with complex structures and non-uniform degradation.}
    \label{fig:example3_full}
\end{figure*}

These examples provide a comprehensive view of TIDE's internal workings across diverse underwater scenes. For each example, we present: (1) the progressive improvement from initial to final restoration with quantitative metrics, (2) the complete restoration pipeline including specialized hypotheses, and (3) the degradation and residual maps that guide our restoration process.

In Figure~\ref{fig:example1_full}, we analyze a diver scene with significant greenish color cast. The progressive improvement visualization shows a +0.24dB PSNR gain from the refinement stage, with noticeable enhancements in the diver's equipment details and seafloor texture. The hypotheses visualization reveals how different decoders contribute to the restoration: the color decoder effectively corrects the greenish tint, while the contrast and detail decoders enhance visibility of the diver's equipment. The degradation maps highlight that color distortion is more severe in the upper region of the image, while the residual maps show that refinement focuses primarily on the diver and foreground elements.

Figure~\ref{fig:example2_full} examines a scene with complex coral formations and a diver. This example demonstrates a more substantial refinement improvement of +0.88dB PSNR and +0.0151 SSIM. The hypotheses visualization shows the complementary nature of our specialized decoders: while the color decoder establishes the overall color balance, the detail decoder is crucial for recovering the fine textures in the coral formations. The degradation maps indicate stronger color distortion in the water column and detail loss in the coral region, while the residual maps show targeted refinement in these areas.

Figure~\ref{fig:example3_full} showcases the underwater statue example, which exhibits the most significant refinement improvement among our examples (+1.40dB PSNR). The initial restoration successfully corrects the dominant blue-green color cast, but the refinement stage substantially enhances the statue's textural details and fine structures. The degradation maps reveal stronger color distortion and detail loss concentrated around the statue, while the hypotheses visualization shows that the color and contrast decoders contribute most significantly to the overall restoration in this case.

These examples collectively highlight several key strengths of our approach. First, the progressive refinement stage consistently improves restoration quality, with PSNR gains ranging from 0.24 dB to 1.40 dB across different cases. Second, the degradation and residual maps reveal the model’s ability to detect and adapt to spatially varying degradation patterns, enabling localized, context-aware adjustments. Third, the specialized decoders produce complementary hypotheses that target different degradation types, which are then effectively fused through adaptive weighting. Finally, the residual maps guide the refinement stage to selectively focus on areas that require further enhancement, while preserving regions that have already been satisfactorily restored.

The internal visualizations provided here offer insights into why our approach outperforms existing methods, particularly on challenging scenes with non-uniform degradation characteristics. By explicitly modeling different degradation types and their spatial distribution, TIDE achieves more targeted and effective restoration compared to conventional approaches that apply uniform processing across the entire image.

\end{document}